\documentclass[nohyperref]{article}

\usepackage{microtype}
\usepackage{graphicx}
\usepackage{subfigure}
\usepackage{booktabs} % for professional tables

\usepackage{hyperref}

\usepackage[accepted]{icml2023}

\usepackage{amsmath}
\usepackage{amssymb}
\usepackage{mathtools}
\usepackage{amsthm}

\usepackage[capitalize,noabbrev]{cleveref}

\theoremstyle{plain}

\theoremstyle{definition}

\theoremstyle{remark}

\let\bs\boldsymbol
\let\ol\overline
\usepackage{physics} %derivatives

\newcommand*{\Scale}[2][4]{\scalebox{#1}{$#2$}}%

\usepackage[textsize=tiny]{todonotes}

\icmltitlerunning{Linear CNNs Discover the Statistical Structure of the Dataset Using Only the Most Dominant Frequencies}

\begin{document}

\twocolumn[
\icmltitle{Linear CNNs Discover the Statistical Structure of the Dataset \\ Using Only the Most Dominant Frequencies}



\icmlsetsymbol{equal}{*}

\begin{icmlauthorlist}
\icmlauthor{Hannah Pinson}{yyy}
\icmlauthor{Joeri Lenaerts}{yyy}
\icmlauthor{Vincent Ginis}{yyy,xxx}
\end{icmlauthorlist}

\icmlaffiliation{yyy}{Data Lab/Applied Physics, Vrije
Universiteit Brussel, Pleinlaan 2, 1050 Brussel, Belgium}
\icmlaffiliation{xxx}{Harvard John A. Paulson
School of Engineering and Applied Sciences, Harvard University, 29
Oxford Street, Cambridge, Massachusetts 02138, USA}

\icmlcorrespondingauthor{Hannah Pinson}{hannah.pinson@vub.be}

\icmlkeywords{Machine Learning, ICML}

\vskip 0.3in
]



\printAffiliationsAndNotice{}  % leave blank if no need to mention equal contribution

% \begin{abstract}
% Our theoretical understanding of the inner workings of convolutional neural networks (CNN) is limited. We here present a new stepping stone towards such understanding in the form of a theory of learning in linear CNNs. By analyzing the gradient descent equations, we discover that using convolutions leads to a mismatch between the dataset structure and the network structure. We show that linear CNNs discover the statistical structure of the dataset with non-linear, ordered, stage-like transitions, and that the speed of discovery changes depending on this structural mismatch. Moreover, we find that the mismatch lies at the heart of what we call the ``dominant frequency bias'', where linear CNNs arrive at these discoveries using only the dominant frequencies of the different structural parts present in the dataset. We furthermore provide experiments that show how our theory relates to deep, non-linear CNNs used in practice. Our findings can help explain several characteristics of general CNNs, such as their shortcut learning and their tendency to rely on texture instead of shape. 
% \end{abstract}

\begin{abstract}
We here present a stepping stone towards a deeper understanding of convolutional neural networks (CNNs) in the form of a theory of learning in linear CNNs. Through analyzing the gradient descent equations, we discover that the evolution of the network during training is determined by the interplay between the dataset structure and the convolutional network structure. We show that linear CNNs discover the statistical structure of the dataset with non-linear, ordered, stage-like transitions, and that the speed of discovery changes depending on the relationship between the dataset and the convolutional network structure. Moreover, we find that this interplay lies at the heart of what we call the ``dominant frequency bias'', where linear CNNs arrive at these discoveries using only the dominant frequencies of the different structural parts present in the dataset. We furthermore provide experiments that show how our theory relates to deep, non-linear CNNs used in practice. Our findings shed new light on the inner working of CNNs, and can help explain their shortcut learning and their tendency to rely on texture instead of shape. 
\end{abstract}

In addition to a neural network's pre-defined architecture, the parameters of the network obtain an \emph{implicit} structure during training. For example, it has been shown that weight matrices can exhibit structural patterns, such as clusters and branches \cite{voss2021branch, casper2022graphical}. On the other hand, the input dataset also has an implicit structure arising from patterns and relationships between the samples. E.g., in a classification task, dogs are more visually similar to cats than to cars. A general theory on how the implicit structure in the network arises and how it depends on the structure of the dataset has yet to be developed. Here we derive such a theory for the specific case of two-layer, linear CNNs, and we provide experiments that show how our insights relate to the evolution of learning in deep, non-linear CNNs. Our approach is inspired by previous work on the learning dynamics in linear fully connected neural networks (FCNN) \cite{saxe2014exact, saxe2019mathematical}, but we uncover the role of convolutions and show how they fundamentally alter the internal dynamics of learning. We start by discussing the two involved structures in terms of singular value decompositions (SVD): on the one hand the SVD of the input-output correlation matrix, representing the statistical dataset structure (Sec. \ref{sec:structure_dataset}); and the SVD of the convolutional network structure on the other hand (Sec. \ref{sec:network_structure}). We subsequently consider the equations of gradient descent in the slow learning regime, yielding a set of differential equations (Sec. \ref{sec:GD_equations}). These equations describe the evolution of the implicit network structure, given the statistical dataset structure. Our analysis reveals that the convolutional network structure gives rise to additional factors in those gradient descent equations: these factors represent the interplay between the singular vectors associated to the dataset and those associated to the convolutional network (Sec. \ref{sec:compare_FCNN}). This interplay changes the speed of discovery of the different parts of the dataset structure, i.e., the singular vectors representing broader to finer distinctions between classes, with respect to the speed of discovery in a FCNN  (Sec. \ref{sec:learning_dynamics}). Internally, this interplay also leads to a dominant frequency bias: only the dominant frequency components of each singular vector associated to the dataset are used by the CNN (Sec. \ref{sec:dominant_frequency_bias}). Experiments with more general datasets confirm the overall dynamics of learning and the existence of the dominant frequency bias (Sec. \ref{sec:experiments}). Finally, we show how our theory relates to deep, non-linear CNNs used in practice (Sec. \ref{sec:nonlinear}). \\ 
Our results can be put in the context of the implicit regularisation resulting from training with gradient descent 
%Neural networks used in practice contain much more trainable weights than there are input samples. Therefore, the target mapping between input and output could potentially be realised by a lot of different weight settings. However, using gradient descent yields specific outcomes for the (evolution of the) network weights
~\cite{du2019gradient, arora2019fine, gidel2019implicit, advani2020high, satpathi2021dynamics}. In \cite{saxe2014exact, saxe2019mathematical} the authors show that the structure of the final weights of linear FCNNs reflect the dataset structure, at least when starting from random initial conditions and when using gradient descent. They find analytical solutions for the learning dynamics in a two-layer, linear FCNN. These solutions indicate the stage-like discovery of different structural parts of the dataset. In \cite{gidel2019implicit}, the discrete dynamics are studied and in \cite{braun2022exact, atanasov2022neural}, the authors develop the theory for different initialisation regimes. The relationship with a.o. early stopping and learning order constancy for both CNNs and FCNNs is further studied  in \cite{hacohen2022principal} as the principal component bias or PC-bias. Furthermore, similar theories exist for linear and shallow non-linear auto-encoders \cite{bourlard1988auto, pretorius2018learning, refinetti2022dynamicsICML}. Finally, it has been shown that linear CNNs trained with gradient descent exhibit an implicit bias in the form of sparsity in the frequency domain \cite{gunasekar2018implicit}. We here show which mechanism gives rise to this sparsity in the frequency domain, and we find \emph{which} frequencies are developed over time. 

\section{Prerequisites}
\textbf{Notation} The input consists of  $n \times n$ images denoted $\bs{X}^{s}$ where $s$ is the sample index. We will omit this index when the context is clear. Often, we will need a vectorized (or `flattened') representation of  $\bs{X}$ and all other $n \times n$ matrices: we turn those matrices in $n^2 \times 1$ vectors through stacking all rows one after the other, and transposing the result. The resulting column vector is denoted with a lower case letter, e.g., $vec(\bs{X}) = \bs{x}$. We reverse this operation to show the vectors as 2D images in figures. An index $j$ into the vector which results from vectorizing a 2D matrix, denoted a `vec-2D' index, runs from $0$ to $n^2-1$. The corresponding indices in the 2D matrix are given by $l = div(j,n)$ and $m = mod(j,n)$, with $div$ integer division and $mod$ the modulo operator. The row $l$ of a matrix $\bs{B}$ is denoted $\bs{B}_{l,:}$ ; the column $m$ is denoted $\bs{B}_{:,m}$. $*$ denotes the complex conjugate. 

\phantomsection  %such that we can link to this specific part
\label{sec:assumptions} 
\textbf{Architecture and assumptions} In our theory, we consider a CNN with a single convolutional layer, a flatten layer, and a single fully connected layer. There are only linear activation functions. The convolutional layer consists of a single kernel $\bs{K}$ with the same dimensions as the input images ($n \times n$). When the kernel reaches the boundaries of the image, it  `wraps' around, such that the convolution is circular. This simplifies the math while still being a good approximation to the zero-padding often used in practice \cite{gray2006toeplitz}. The task is image classification with one-hot encoded labels, we assume a slow learning rate, an MSE loss, and small, random initial conditions. 

\phantomsection  %such that we can link to this specific part
\label{sec:SVD} 
\textbf{The singular value decomposition (SVD)}
The SVD of an $p \times n^2$ ($p < n^2$) real-valued matrix $\bs{B}$ is given by: $\bs{B}  = \bs{U S V}^{T}$, where $\bs U$ is an orthogonal $p \times p$ matrix  ($\bs{U}^{T} = \bs{U}^{-1}$) containing the left singular vectors as columns, $\bs V$ is an orthogonal $n^2 \times n^2$ matrix containing the right singular vectors as columns, and $\bs S$ is a $p \times n^2$ rectangular diagonal matrix containing the $p$ real and positive singular values, denoted $s_\alpha$. The singular vectors and values are by definition sorted from highest to lowest singular value. The modes $\bs{M}^{(\alpha)}$, $\alpha \in \{0, \dots, p\}$ form the decomposition of the original matrix: $\bs{B}  = \sum_\alpha \bs M^{(\alpha)}$, with $\bs M^{(\alpha)} = s_\alpha \, \bs U_{:, \alpha} \bs V^{T}_{\alpha,:}$. 

\begin{figure}[h!]
   \centering
   \includegraphics[width=0.45\textwidth]{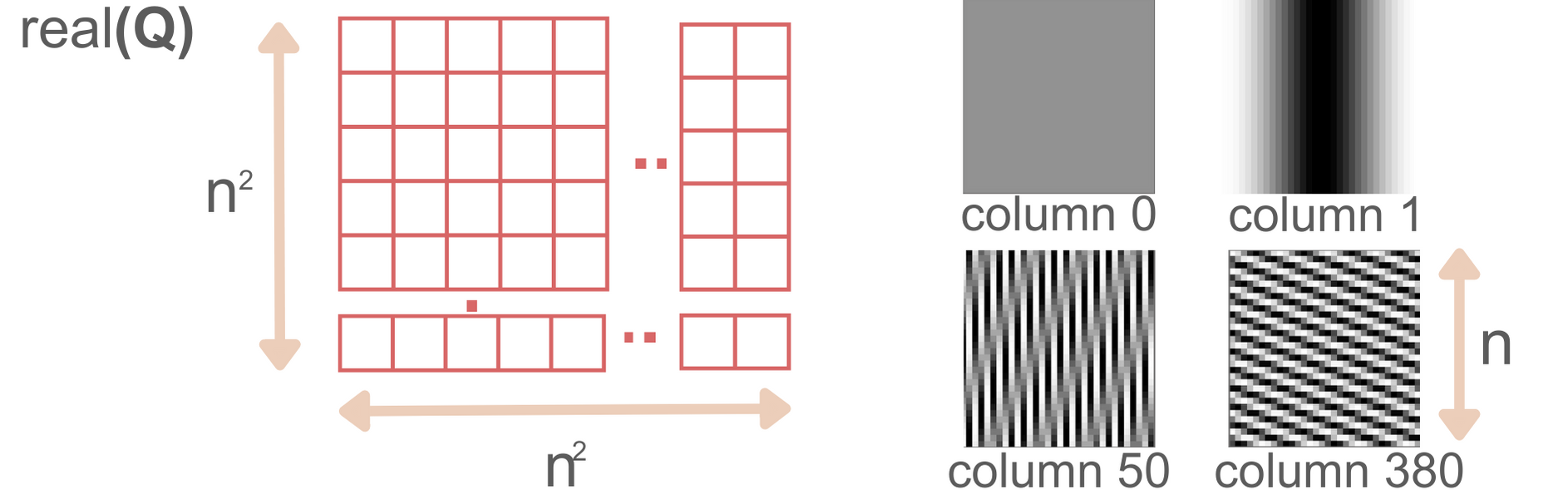}
   % \includegraphics[width=0.45\textwidth]{ICML_final_figures/Q.png}
   %\captionof{figure}{Figure caption}
   \caption{ The symmetric matrix $\bs{Q}$ of wich each row/column is a vectorized Fourier eigenmatrix (4 examples given). }
   \label{fig:real_Q}
 \end{figure}

\phantomsection  %such that we can link to this specific part
\label{sec:vec_2D_DFT} 
\textbf{The vectorized 2D discrete Fourier transform} The Fourier transform is another type of decomposition, used to decompose a signal in its frequency components. The resulting amplitudes in the frequency domain, the `Fourier coefficients', describe the relative importance of each frequency in the original signal. For an $n \times n$ matrix $\bs{B}$, the conventional 2D discrete Fourier transform (denoted $\mathfrak{F}_{2D}$) is given by $\mathfrak{F}_{2D}(\bs{B})_{\mu,\nu} = \Sigma^n _{s,t}   (\omega_n)^{\mu s}  (\omega_n)^{\nu t} B_{s,t}$ with $\omega_n = \exp(-2 \pi i / n)$ and $i^2 = -1$. The result is a 2D $n \times n$ complex-valued matrix, whose values tell us the relative importance of the $n^2$  different spatial frequencies: $n$ frequencies in the vertical direction, combined with $n$ frequencies in the horizontal direction. We will make use of an equivalent transform, but instead of applied to a 2D matrix, applied to the 1D vector that results from vectorizing the $n \times n$ matrix first. This transform is given by multiplication with the $n^2 \times n^2$ matrix $\bs{Q}$, i.e., $\bs{Q} \bs{x} = \frac{1}{n}vec(\,  \mathfrak{F}_{2D}(\bs X) \, )$. $\bs{Q}$ is illustrated in Fig.\ref{fig:real_Q}; for its exact definition, see \hyperref[sec_sup:Q]{SI}. Note that this operation is different from the 1D discrete Fourier transform of a vector: we will call it the vectorized 2D discrete Fourier transform, or vec-2D DFT in short.  

If the original vector is real-valued, the Fourier spectrum of this vector will exhibit symmetries. Let a vec-2D frequency index $j$ correspond to a pair of horizontal and vertical frequency indices $(\mu, \nu)$, then the symmetric vec-2D frequency index $j_{symm}$ corresponds to the pair $(n-\mu, n-\nu)$, thus $j =  (n-1) \mu + \nu$ and $j_{symm} = (n-1) (n-\mu) + (n-\nu)$. The symmetry in the spectrum is then given by the equation $(\bs{Q} \bs{x})_{j_{symm}} = (\bs{Q} \bs{x})^*_{j}$.

\begin{figure*}
\centering
 \includegraphics[width=\textwidth]{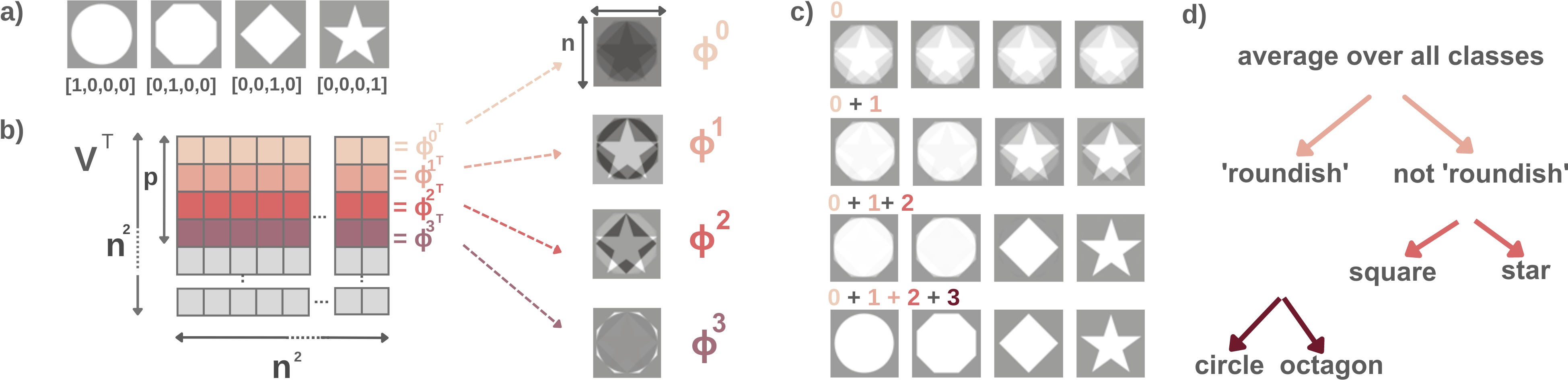}
\caption{Illustration of the relationship between statistical structure and SVD. $\bs{a)}$ Example dataset consisting of geometric shapes: a circle, an octagon, a square and a star. There is only one sample per class. $\bs{b)}$ Visualisation of $\bs{V}^T$ and its first p rows, the singular vectors denoted $\bs{\phi}^{\alpha}$ (here p=4), as $n \times n$ matrices. The other rows carry no meaning. $\bs{c)}$ Visualisation of the summation of the modes $\bs{M}^{(\alpha)}$ of $\bs{\Sigma^{ y  x}}$, given by the SVD (see \hyperref[sec:SVD]{prerequisites}). Each mode is first reshaped to p $n \times n$ matrices.  $\bs{d)}$ Interpretation of the implicit hierarchical structure in $\bs{\Sigma^{ y  x}}$, which can be derived from the visualisations in subfigures b) and c).} %$\bs{e)}$ Vec-2D Fourier coefficients of the 4 singular vectors, i.e., $|\bs{Q} \bs{\phi}^{\alpha}|_j$ in function of the vec-2D indices $j$. The values for $j > 2048$ are given by the symmetry properties of Fourier transforms. When comparing with subfigure b), we see that broader distinctions in shape correspond to spectra with limited peaks at low frequencies, while fine distinctions also contain higher frequencies. }
\label{fig:geometric_dataset_structure}
\end{figure*}

\section{Differential Equations of Gradient Descent}
    
We want to study the relationship between the dataset structure and the network structure during training with gradient descent. To this end, we first discuss the SVDs that capture these structures. We will then show how the equations of gradient descent can be reformulated in terms of those SVDs.

\subsection{Statistical Structure of the Dataset}\label{sec:structure_dataset}

The dataset has an implicit structure, e.g., some images contain the same backgrounds, and images of cats are similar to images of dogs. Given a general dataset, it is unclear a priori which patterns might be relevant during the training of a network.  For linear CNNs, however, we find that the relevant structure is given by the statistical structure: the structure that can be derived from the correlations in the dataset. In Fig.\ref{fig:geometric_dataset_structure}, we illustrate the statistical structure of a dataset consisting of geometric shapes. 
Given ground-truth labels $\bs{y}$ when there are $p$ classes, we can define the $p \times n^2$ input-output correlation matrix $\bs{\Sigma^{ y  x}} $ as: 
\begin{equation} \label{eq:sigma_yx}
\bs{\Sigma^{ y  x}} = \langle  \bs{y}  \bs {x}^T \rangle,
\end{equation}
where $\bs{x}$ is an $n^2 \times 1$ vectorized sample, and $\langle \rangle$ denotes the average over all samples. Note that $\bs{y}  \bs {x}^T$ is an outer product, i.e., $\Sigma^{ y  x}_{l,m} = \langle  y_l  x_m \rangle$. Given one-hot encoded labels, each row of $\bs{\Sigma^{ y  x}}$ contains the vectorized average image of each class. 
The structure relevant when training a (linear) neural network is captured through the singular value decompositions (SVD) of this matrix (see \cite{saxe2019mathematical} and Fig. \ref{fig:geometric_dataset_structure}(b-d)). The SVD of $\bs{\Sigma^{ y  x}}$ is given by:
\begin{equation}
\label{eq:SVD_sigma_y_x}
\bs{\Sigma^{ y  x}}  = \bs{U S V}^{T}.
\end{equation}
In this particular case, the first $p$ rows of $\bs{V}^T$, which we denote $\bs{\phi}^{\alpha}$, $\alpha \in \{0, \dots, p-1\}$, are the principal components of the 
averaged class images \cite{pearson1901liii, jolliffe2002principal}. The $p \times p$ matrix $\bs{U}$ then contains the coefficients to reconstruct the $p$ vectorized, averaged class images as linear combinations of the $p$ right singular vectors $\bs{\phi}^{\alpha}$. By visualizing the singular vectors $\bs{\phi}^{\alpha}$ and the corresponding modes $\bs{M}^{\alpha}$, we can see the SVD here embodies broader to finer distinctions between the averaged class images (see Fig.\ref{fig:geometric_dataset_structure}(b-d)). As such, it captures our intuitive understanding of an implicit hierarchical structure between the shapes in the dataset. \\
We can also consider the singular value decomposition of the input-input correlation matrix $\bs{\Sigma^{ x x}} = \langle  \bs{x}  \bs {x}^T \rangle$. In general, this matrix could have $n^2$ different singular values and corresponding singular vectors, and its exact shape influences the dynamics of learning as well. However, to focus on the effect of a convolutional architecture only, we consider the case where $\bs{\Sigma^{ x x}}$ is diagonalizable in the basis given by $\bs{V}$:
\begin{equation}
    \label{eq:sigma_x_x_V}
\bs{\Sigma^{ x x}}= \bs{V \overline{\Sigma^{ x x}} V^{T}}~,
\end{equation}
where $\bs{\overline{\Sigma^{ x x}} }$ is a diagonal matrix with only the first $p$ entries on the diagonal non-zero. In our case of classification with one-hot encoded labels, this is the case if all images in a class are the same (see \hyperref[sec_sup:sigma_x_x]{SI}). The less images deviate from the average image for their class, the better this assumption holds. %In the section 'experimental results', we show results for the CIFAR-10 dataset where this assumption does not hold exactly; however, previous study has shown that for most standard datasets used in image classification, the assumption holds well (REF).  \textcolor{red}{discuss relationship to overfitting}
Finally, given any set of predictions $\bs{\hat y}$ for a set of input samples $\bs{x}$, we can also compute the input-predicted output correlation matrix $\bs{\Sigma^{\hat y  x}}$ (cfr. Eq. \ref{eq:sigma_yx}):
\begin{equation}
\label{eq:sigma_y_hat_x}
\bs{\Sigma^{\hat y  x}}  = \langle \bs{ \hat y  x^T } \rangle,
\end{equation}
where $\bs{\Sigma^{\hat y  x}}$ has dimensions $p \times n^2$. 
% In the next sections,  we will compare the evolution of this matrix between linear FCNN and linear ConvNN, and will subsequently explain how the difference in evolution arises from the constrained network structure of the ConvNN. At the same time, the analysis will show how the structure of the dataset (eq. \ref{eq:SVD_sigma_y_x}) is discovered by the linear ConvNN. Given differential equations involving $\bs{k}$, $\bs{W}$, and the different correlation matrices, we can reveal the interaction between the different structures through performing a change of coordinates.
To relate the evolution of this matrix to the structure of the dataset, we study $\bs{\Sigma^{\hat y  x}}$ in the SVD basis associated to $\bs{\Sigma^{ y  x}} $ (see Eq. \ref{eq:SVD_sigma_y_x}):
\begin{equation}
\label{eq:SVD_sigma_y_hat_x}
\bs{\Sigma^{\hat y x}}  = \bs{ U A  V}^{T}.
\end{equation}
The $p \times n^2$ matrix $\bs{A}$ is not necessarily rectangular diagonal; if it would be, and if the diagonal elements would be positive real numbers, Eq. \ref{eq:SVD_sigma_y_hat_x} would be an SVD with singular values given by the values on the diagonal of $\bs{A}$. In that case, we could say that $\bs{\Sigma^{\hat y  x}}$ has the same structural elements as $\bs{\Sigma^{ y  x}}$, only with different strengths.

\subsection{Structure of the Network}\label{sec:network_structure}
%Furthermore, the convolution is really a convolution (thus using a flipped kernel) and not a correlation, as it is often implemented in practice (REF). This has no real influence other than simplifying the math (see \textcolor{red}{SI}).

% \begin{minipage}{\linewidth}
%    \centering
%    \includegraphics[width=0.7\linewidth]{figures/draft_real_Q.png}
%    %\captionof{figure}{Figure caption}
%    \captionof{figure}{ Illustration of the eigenvectors of the convolutional layer with a single kernal. The matrix  $\bs{Q}$ has the $n^2$ vectorized eigenmatrices of the 2D discrete Fourier transform as $1 \times n^2$ columns. We illustrate the real part of 4 of the $n^2$ columns as $n \times n$ matrices. }
%    \label{fig:real_Q}
%  \end{minipage}

We also want to study the implicit structure of the convolutional network through an SVD. To this end, we replace the convolution operations with convolution-equivalent weight matrices, such that we can study the SVD of these matrices \cite{sedghi2019singular}. A convolutional layer can be seen as a constrained fully connected layer: the sliding of the kernel over the image is actually a repeated application of the same weights \cite{Goodfellow-et-al-2016}. A convolution-equivalent weight matrix is then a matrix where elements are repeated in a fixed, circulant pattern. Such a matrix is called a doubly block circulant matrix (see  \citealt{jain1989fundamentals, sedghi2019singular}), denoted $\bs{dbc}$ (definition, see \hyperref[sec_sup:dbc]{SI}). We can also define such a matrix through its eigendecomposition: the  $n^2 \times n^2$ doubly block circulant matrix of the kernel $\bs{K}$ is given by
\begin{equation}
    \bs{dbc}(\bs{K}) =  n \, \bs{Q}^{-1}  diag(\bs{Q k} ) \bs{Q}
    \label{eq:dbc_k_eigendecomp}
\end{equation}
and it thus acts as a convolution-equivalent weight matrix:
\begin{equation}
   vec(\bs{X} \circledast \bs{K} ) = \bs{dbc}(\bs{K}) \bs x
   \label{eq:convolution-equivalent}
\end{equation}
where $\circledast$ denotes circular convolution. Here $\bs{k}$ is the vectorized kernel, and $\bs{Q}$ is the matrix corresponding to the vec-2D DFT (see \hyperref[sec:vec_2D_DFT]{prerequisites} and Fig. \ref{fig:real_Q}). $diag(\bs{Q k} )$ is an $n^2 \times n^2$ diagonal matrix with the $n^2$ elements of the vec-2D DFT of $\bs{k}$, thus $\bs{Qk}$, on the diagonal. Readers familiar with the convolution theorem might recognize it in Eq. \ref{eq:dbc_k_eigendecomp}: to apply a convolution, transform the input to the Fourier domain, multiply with the transformed kernel, and transform the result back to the original domain. \\
In short, for our theoretical derivations, we consider the network with predictions $\bs{\hat y}$ given by:
\begin{equation}
\label{eq:y_hat}
   \bs{ \hat y}  =  \bs{W} \bs{dbc(K)} \bs{x}, 
\end{equation}
with $\bs{W}$ the $p \times n^2$ weight matrix of the fully connected layer. We can again formalize the notion of implicit structure through considering the singular value decompositions/eigenvalue decompositions of the network's weight matrices  $\bs{W}$ and  $\bs{dbc(K)}$. While a general weight matrix $\bs{W}$ could in principle have any singular-/eigenvalue decomposition, reflecting its unconstrained structure, the eigendecomposition of $\bs{dbc(K)}$ is \emph{fixed} and always given by Eq. \ref{eq:dbc_k_eigendecomp}: the eigenvectors stay the same. Only the eigenvalues, given by $n \bs{Q k}$, can change. This reflects the fact that of the $n^2 \times n^2$ matrix  $\bs{dbc(K)}$, only $n^2$ values can change independently. These values are given by the $n \times n$ kernel. The different singular values of $\bs{dbc(K)}$ are actually given by $n |(\bs{Q k})_i|$. For a discussion on the role of the phases, see \hyperref[sec:lagrange_sol]{SI}.

\subsection{Differential Equations of Gradient Descent}\label{sec:GD_equations}
We can now focus on what happens during training with gradient descent. We start from a MSE loss function $ L = \frac{1}{2} \sum_{l=0}^p (y_l - \hat y_l)^2$, and its gradient is used to update the network parameters at every discrete timestep:
\begin{align}
    & \Delta \bs k =  - \lambda  \frac{\partial L }{\partial \bs k^T}, 
     \label{eq:discrete_update_k}
   & \Delta \bs W = - \lambda  \frac{\partial L }{\partial \bs W^T},
   %\label{eq:discrete_update_W}
\end{align}
with $\lambda$ the learning rate. For the convolutional layer, the gradient of the loss with respect to the kernel is in itself
given by a convolution, but with a flipped image  (see, e.g., \citealt{Goodfellow-et-al-2016}, Ch. 9). The gradients are then given by:
% \begin{align}
%      \frac{\partial L }{\partial \bs k^T} 
%      &= vec \big ( \bs{X}_{flip} \ast \frac{\partial L }{\partial \bs H^T} \big ) = dbc(\bs{X}_{flip}) \sum_{l=0}^p (y_l - \hat y_l)  (\bs W^T)_{:,l}.
% \end{align}
\begin{align}
    & \frac{\partial L }{\partial \bs k^T} 
   = dbc(\bs{X}_{flip}) \sum_{l=0}^p (y_l - \hat y_l)  (\bs W^T)_{:,l} \\
    & \big(\frac{\partial L }{\partial \bs W^T}\big)_{l,:}  =  (y_l - \hat y_l) \bs{x}^T dbc(\bs k)^T .
\end{align}
In the slow learning regime, the weights and kernel change minimally with each update. The updates can then be approximated through an averaged update over samples (\citealt{saxe2019mathematical}, SI p1.). This allows us to introduce the matrices $\bs{\Sigma^{ y  x}}$ (cfr. Eq. \ref{eq:sigma_yx}) and $\bs{\Sigma^{\hat y x}}$ (cfr. Eq. \ref{eq:SVD_sigma_y_hat_x}) in the equations. Given the slow learning rate, the discrete equations can also be reformulated as differential equations. We can subsequently transform the different variables using the SVD basis of $\bs{\Sigma^{ y  x}}$:
\begin{align}
     \label{eq:transform_W}
 \bs{\ol{W}} & = \bs{ U^{T}} \bs W \bs{R}  \\
    \label{eq:transform_dbc_K}
\ol{\bs{dbc(K)}} & = \bs{R}^{-1} \bs{dbc(K)}  \bs{V}
\\ &= n \bs{R}^{-1}  \bs{Q^{-1}} diag(\bs{Qk})  \bs{Q}  \bs{V} 
\end{align}
here $\bs{R}$ is an arbitrary $n^2 \times n^2$ invertible matrix. This matrix reflects the freedom in the actual shape of the weights, as long as their product remains the same.  Now we have $ \Scale[0.9]{\bs{\Sigma^{\hat y x}}= \bs{W} \bs{dbc(K)} \bs{\Sigma^{ x x}} =  \bs{U} \bs{A} \bs{V}^{T}  = \bs{ U \, \ol{W} \, \ol{\bs{dbc(K)}}  \, \ol{\Sigma^{ x x}} \, V}^{T}}$ (see also Eq. \ref{eq:sigma_x_x_V}), and we arrive at the differential equations in the following, somewhat complicated shape (for full derivation, see \hyperref[sec_sup:diff_eq]{SI}) :
\begin{align}
\label{eq:update_diag_Qk_transformed}
& \Scale[0.85]{\frac{1}{n \lambda}  \big ( \frac{d \, diag(\bs{Q^{-1} k})}{dt}  \big )_{j,j} = \bs Q^{-1}_{j,:} \bs{R}^{-*T} \bs{\ol{W}} ^{*T}   \big( \bs{S} - \bs{A}     )  (\bs{Q} \bs{V} )^T_{:,j},} \\
\label{eq:update_W_bar}
& \Scale[1]{\frac{1}{n\lambda}  \big( \frac{d \bs{\ol{W}} }{dt}  \big) =   \big( \bs{S} - \bs{A}     )  (\bs{Q} \bs{V} )^T diag(\bs{ Q k})  \bs{Q}^{-1} \bs{R}},
\end{align}
where $diag(\bs{Q^{-1} k})_{j,j} = (\bs{Q^{-1}k})_j = (\bs{Q^{*T}k})_j$, and all derivatives of off-diagonal elements of $diag(\bs{Q^{-1} k})$ are zero.
There are a couple of key observation to make about this set of differential equations (Eqs. \ref{eq:update_diag_Qk_transformed} and \ref{eq:update_W_bar} ): firstly, they are coupled and non-linear, making them difficult or impossible to solve analytically but for very specific cases. We can also see that at every step $\bs{A}$ is brought closer to $\bs{S}$  (cfr. Eq. \ref{eq:SVD_sigma_y_x}), that $\bs{A}$ thus becomes diagonal, and that the updates become zero if these matrices are equal: i.e., the predictions perfectly match the labels. Furthermore, the dynamics of learning are influenced by the vectors $\bs{Q} \bs{V}_{:, \alpha} = \bs{Q} \bs{\phi^{\alpha}} $, the vec-2D Fourier transforms of the right singular vectors $\bs{\phi^{\alpha}}$. This turns out to be a crucial point: this embodies the interplay between the dataset structure (characterised by $\bs{V}$) and the convolutional network structure (characterised by $\bs{Q}$); in the equivalent equations for fully connected networks, no such extra factors are present (see Eqs. \ref{eq:update_W_1_bar} and \ref{eq:update_W_2_bar} below). 

\subsection{Comparison with Fully Connected networks}  \label{sec:compare_FCNN}

 The dynamics of learning during gradient descent in linear FCNNs are discussed in \cite{saxe2014exact, saxe2019mathematical}. In the case of a two-layer linear FCNN, we have 
\begin{equation}
\Scale[0.90]{
   \bs{ \Sigma^{\hat y x}}_{FC} = < \bs{\hat y}_{FC} \bs{x^T} > =  \bs{W^2W^1 \Sigma^{xx} } = \bs{U} \bs{A}_{FC} \bs{V}^{T}},
\end{equation}
where $\bs{W^2}$ and $\bs{W^1}$ are the weight matrices for the second and first layer of the FCNN network, respectively. The matrices $\bs{U}$ and $\bs{V}$ are the same as before (Eq. \ref{eq:SVD_sigma_y_x}). However, the evolution of the matrices  $\bs{ \Sigma^{\hat y x}}_{FC}$  and $\bs{A}_{FC}$ will be different from the evolution of their equivalents in the CNN case. 
Using transformed weight matrices:
\begin{align}
\label{eq:W1_bar}
  & \bs{ \ol{W^1} } = \bs{R'^{-1}} \bs{W^1} \bs{V}, &\bs{ \ol{W^2} } = \bs{U^{T}} \bs{W^2} \bs{R'},
\end{align}
with $\bs{R'}$ an arbitrary invertible matrix, the authors in \cite{saxe2014exact, saxe2019mathematical} then arrive at the system of differential equations:
\begin{align}
\label{eq:update_W_1_bar}
    \frac{1}{\lambda}  \dv{\bs{\ol{W^1}}}{t} = \bs{\ol{W^2}}^T (\bs{ S - A_{FC} }) \\
\label{eq:update_W_2_bar}
    \frac{1}{\lambda}  \dv{\bs{\ol{W}^2}}{t} = (\bs{ S - A_{FC}}) \bs{\ol{W}^1}^T. 
\end{align}
which can be compared to the equations we derived for the CNNs (Eqs. \ref{eq:update_diag_Qk_transformed} and \ref{eq:update_W_bar} and  \hyperref[sup_sec:align_balanced_WTA]{SI}). The authors subsequently show that, starting from small, random initial conditions,  $\bs{\ol{W}}^{2}$ and $\bs{\ol{W}}^{1}$ each quickly become diagonal. Comparing this to the case of the CNN, we see that $\ol{\bs{dbc(K)}}$ cannot be diagonal (Eq. \ref{eq:transform_dbc_K}). This indicates that the constrained structure of the convolutional layer in itself cannot fully reflect the dataset structure, while a fully connected layer in an FCNN can. 

% % Furthermore, when comparing these equations to their counterparts for a single layer, linear, fully connected neural network (derived in \cite{saxe2019mathematical}, see also supplemental), we 

% % \begin{align}
% % \label{eq:update_W_1_bar}
% %     \frac{1}{\lambda}  \dv{\bs{\ol{W^1}}}{t} = \bs{\ol{W^2}}^T (\bs{ S - A }) \\
% % \label{eq:update_W_2_bar}
% %     \frac{1}{\lambda}  \dv{\bs{\ol{W}^2}}{t} = (\bs{ S - A}) \bs{\ol{W}^1}^T. 
% % \end{align}
% % where $\bs{\ol{W^1}}$ and  $\bs{\ol{W^2}}$ are the transformed versions of the first and second weight matrix, respectively. 

\section{Non-linear Learning Dynamics} \label{sec:learning_dynamics}
\begin{figure}[h!]
   \centering\includegraphics[width=0.47\textwidth]{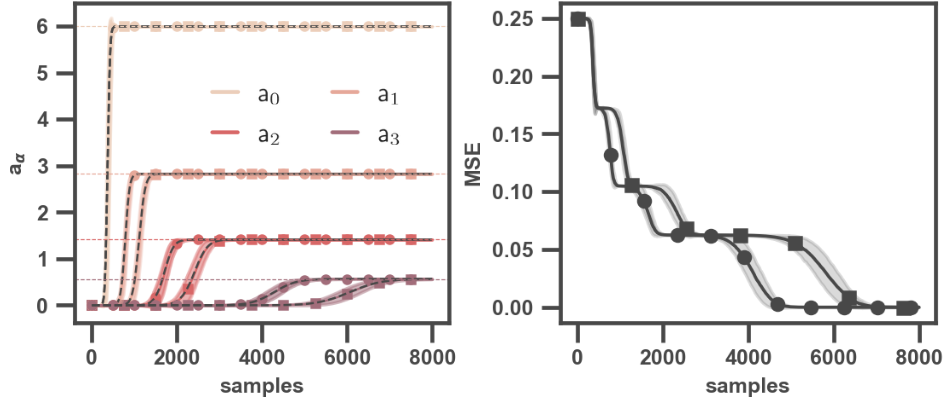}
   %\captionof{figure}{Figure caption}
   \caption{Experimental evolution and theoretical predictions (left) of $a_\alpha$, and experimental evolution of the MSE loss (right), for linear FCNN (round markers) and linear CNN (square markers), trained on the same dataset. The black, dashed lines indicate the predictions. All plotted lines are averaged over trials, shadowed regions indicate the standard deviations. Dashed horizontal lines left indicate the values $s_\alpha$.} 
   \label{fig:analytical_results}
 \end{figure}

We now explore how the interplay between $\bs{V}$ and $\bs{Q}$ influences the dynamics of learning. In particular, we study the evolution of the predictions: we can study this evolution through analysing the evolution of the matrix $\bs{A}$ (see Eq. \ref{eq:SVD_sigma_y_hat_x}), with $ \bs{A} = \bs{\ol{W}} \, \ol{\bs{dbc(K)}}  \, \bs{\ol{\Sigma^{ x x}}}$.  We will first derive analytical solutions for this evolution for a particular, illuminating dataset. Later we will show that the characteristics of the found trajectories roughly hold for more general datasets as well. 
Consider the dataset where the image for each class is given by:
\begin{equation}\label{eq:pure_cos_input}
    X^{(c)}_{l,m} =  b^{(c)} \cos( 2 \pi \frac{\mu l}{n} +   2 \pi \frac{\nu m}{n}).
\end{equation}
Here $c$ is the class index, and for each class we pick a different pair of  $\mu$ and $\nu$ $\in \{0, \cdots, n-1 \}$, indexing the $n$ possible horizontal and vertical frequencies. $l$ and $m$ index the pixels. $b^{(c)}$ is the amplitude of the frequency (one value per class). Examples of this type of input are shown in Fig. \ref{fig:real_Q}, since the vectorized input images each correspond to the real part of a column of $\bs{Q}$ (apart from the amplitude). Since the class vectors are already orthogonal, normalisation yields the $p$ singular vectors $\bs{V}_{:,\alpha} = \bs{\phi}^{\alpha}$.   This type of input decouples the differential equations (Eqs. \ref{eq:update_diag_Qk_transformed} and \ref{eq:update_W_bar}): the intuition behind this is that the structural  `mismatch' between dataset and network partly vanishes, because we pick the vectors $\bs{\phi}^{\alpha}$ to be orthogonal to the real part of the columns of $\bs{Q}$. However, $\bs{Q}$ is complex-valued and the singular vectors are real-valued; therefore an artefact of the mismatch will remain in the form of a factor $\frac{1}{\sqrt{2}}$. In the \hyperref[sup:analytical_solutions]{SI}, we first define a set of specific initial conditions such that $ \bs{A}$ starts out rectangular diagonal. Subsequently we show that given the Eqs.  \ref{eq:update_diag_Qk_transformed} and \ref{eq:update_W_bar}, each $a_\alpha = \bs{A}_{\alpha, \alpha}$ exactly follows a sigmoidal trajectory: 
\begin{equation}
    a_{\alpha} (t) = \frac{s_{\alpha} e^{ 2 n \lambda d_\alpha  s_{\alpha} t}}{ e^{ 2 \lambda n d_\alpha s_{\alpha} t} - 1 + s_{\alpha} / a_{\alpha}^{(0)} },
\end{equation}
with $d_\alpha = \frac{1}{\sqrt{2}}$ (unless $\mu=0$ and $\nu=0$, then $d_\alpha = 1$), and $a_{\alpha}^{(0)} = a_{\alpha}(t=0).$ The time, in number of samples, to grow from an initial value $a_{\alpha}^{(0)}  = \epsilon$ with $\epsilon \ll 1$, to a value $a_\alpha = s_\alpha - \epsilon$, is approximately given by $\frac{d_\alpha n}{s_\alpha \lambda}$ (see \citealt{saxe2019mathematical}). Under analogous initial conditions and assumptions, the values $a_\alpha(t)$ in a linear FCNN follow a similar sigmoidal trajectory (see \citealt{saxe2019mathematical}). However, there are no factors $d_\alpha$ and $n$: the number of samples needed to reach convergence for each value $a_\alpha$  is approximately given by $\frac{1}{s_\alpha \lambda}$ for FCNN.\\ Given these analytical results, we can draw the following conclusions: first, given the sigmoidal trajectories, the predictions of the network change in such a way that the different structural modes of $\bs{\Sigma^{y  x}} $  are discovered with rapid, stage-like transitions by both types of networks. This discovery is ordered in time (through the factor $\frac{1}{s_\alpha}$), from highest singular value to lowest singular value, corresponding to the discovery from broader to finer distinctions between the classes. However, the CNN exhibits a different effective learning rate $\lambda_{eff} = n d_\alpha \lambda$ for each mode $\alpha$ w.r.t. the FCNN. The factor $n$ is a speed-up resulting from the convolution with a kernel of dimension $n$; the factor $d_\alpha <=1$ reflects a delay stemming from the mismatch between the dataset structure and the constrained network structure.
In  Fig. \ref{fig:analytical_results}, we show the results of experiments with a dataset of `pure cosines' as described by Eq. \ref{eq:pure_cos_input} (for details, see \hyperref[sec_sup:experiments]{SI}). The first class/singular vector $\bs{\phi^{0}}$ is a constant image, i.e., $\mu = \nu = 0$; the subsequent classes have different randomly selected frequencies with decreasing amplitudes. The FCNN is trained with a higher learning rate $\lambda_{FC} = n \lambda_{CNN} $, such that the graphs of both networks can be compared on the same plot. $d_0 = 1$, such that the trajectories of $a_0$ for the CNN and FCNN overlap. $d_\alpha = 1/\sqrt{2} \approx 0.71$ for the other modes, resulting in a shift between the trajectories. The theoretical predictions, made using the same set of initial conditions, exactly match the experimental trajectories. \\
We derived the analytical solutions using aligned, balanced initial conditions (see \hyperref[sup_sec:align_balanced_WTA]{SI}). These conditions render $\bs{A}$ rectangular diagonal from the beginning. Several previous studies show that when starting from fully random, small initial conditions, $\bs{A}$ very quickly becomes diagonal as well \cite{saxe2014exact, saxe2019mathematical, atanasov2022neural, braun2022exact}. We discuss this further in the \hyperref[sup_sec:align_balanced_WTA]{SI}. Apart from a small additional delay related to this alignment phase, the trajectories of $\bs{A}$ when starting from small, random initial conditions are thus described by similar sigmoidal curves. 

\section{Dominant Frequency Bias} \label{sec:dominant_frequency_bias}

\begin{figure*}[h!]
\centering
\includegraphics[width=\textwidth]{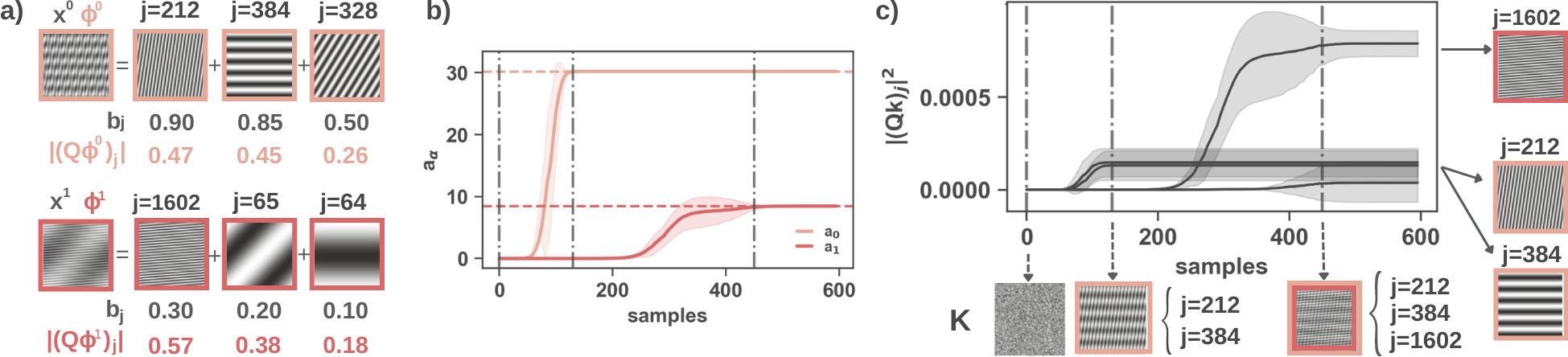}
\caption{ Illustration of the dominant frequency bias. $\bs{a)}$ Illustration of the class images as sums of pure cosines. Since the class vectors $\bs{x}^c$ are orthogonal, the singular vectors $\bs{\phi}^{\alpha}$ are normalised versions of the vectors  $\bs{x}^c$. $\bs{b)}$ Evolution of the effective singular values. $\bs{c)}$ Evolution of the squared Fourier coefficients of the kernel (middle), evolution of the kernel $\bs{K}$ at selected timepoints (below), frequencies corresponding to highest Fourier coefficients of final kernel (right). Compare the development of the kernel over time with the discovery of the modes, shown in b), and the values of  $|\bs{Q} \bs{\phi}^{\alpha}|_j$, shown in a). $\bs{b)}$, $\bs{c)}$: solid lines: averages over experimental runs. Shaded regions: standard deviations. Vertical dashed lines: selected timepoints. Horizontal dashed lines: singular values $s_\alpha$.}
\label{fig:CNN_dominant_freq}
\end{figure*}
So far, we have studied the evolution of the predictions for two-layer linear CNNs and FCNNs. It turns out that the way these networks arrive at those predictions internally is very different. We will show that during training the kernel of the linear CNN becomes an implicit regularizer: it filters out a small fraction of the frequency components present in the dataset. The whole network arrives at its predictions using only those frequencies. Driving this implicit regularisation is a soft winner-takes-all dynamics (sWTA) \cite{lazzaro1988winner,fang1996dynamics,fukai1997simple}, where during training different vec-2D Fourier coefficients of the kernel $|\bs{Q} \bs{k}|_j$ `compete' with each other to be part of the final kernel. Whether they `win' depends on their initial values and the corresponding coefficients of the singular vectors of the dataset $|\bs{Q} \bs{\phi}^{\alpha}|_j$. The word `soft' denotes that there is not a single winning frequency. This sWTA dynamics---and thus the implicit regularisation---can be derived from the given differential equations.\\ 
To see this, we first consider a slightly more general type of dataset: a dataset for which the modes now consist of \emph{sums} of cosines, where each cosine possibly has a different amplitude (see  \hyperref[sup_sec:sums_of_cosines]{SI}).  However, we still assume frequencies are not shared between modes, such that modes remain decoupled. Formally, given an index $j$, $|\bs{Q} \bs{\phi}^{\alpha}|_j$ is non-zero for only one mode $\alpha$.  $\sigma^{(\alpha)}$ then denotes the set of all vec-2D indices $j$ of the frequencies associated to a mode $\alpha$. If $j$ is the vec-2D index corresponding to the pair of indices $(\mu, \nu)$, then with $j_{symm}$ we denote the vec-2D index that corresponds to $(n-\mu, n-\nu)$. Since the singular vectors are real-valued, their Fourier spectra exhibit symmetries, and $|\bs{Q} \bs{\phi}^{\alpha}|_j = |\bs{Q} \bs{\phi}^{\alpha}|_{j_{symm}}$.  $\sigma_{symm}^{(\alpha)}$ includes all the symmetric indices $j_{symm}$ as well.  In the \hyperref[sup_sec:align_balanced_WTA]{SI}, we show that when starting from small, random initial conditions, the development of the vec-2D Fourier coefficients of the kernel $|\bs{Q} \bs{k}|_j$ is approximately given by:
\begin{equation}
\label{eq:diag_winner_takes_all_main}
    \frac{1}{ 2 n \lambda}  \frac{d \,|(\bs{Q k})_j|^2}{dt} =  |(\bs{Q k})_j|^2 \,  |(\bs{Q}\bs{\phi}^{\alpha})_j|  ( s_\alpha- a_\alpha), 
\end{equation}
when $j \in \sigma_{symm}^{(\alpha)} $, with
\begin{align}
\label{eq:a_sum_of_freq_main}
    a_\alpha = & n \ol{\bs{\Sigma}^{ x x}}_{\alpha,\alpha}  \Big( \,2 |(\bs{Q k})_j|^2 |(\bs{Q} \bs{\phi}^{\alpha})_j| \nonumber \\ & + \sum_{j' \in \sigma_{symm}^{(\alpha)}  \setminus  \, \{j, j_{symm}\}}  |(\bs{Q k})_{j'}|^2 |(\bs{Q} \bs{\phi}^{\alpha})_{j'}| \Big).
\end{align}
While $\bs{s}_{\alpha}$ can be considered as the input that drives the system, $- |(\bs{Q k})_j|^2 $ is a self-inhibition term, and $- \sum_{j' \in \sigma_{symm}^{(\alpha)}  \setminus  \, \{j, j_{symm}\}} |(\bs{Q k})_{j'}|^2$ is a term that captures the lateral inhibition coming from the other 2D-vec kernel Fourier coefficients associated to the same mode. A formulation and analysis of sWTA dynamics close to the equations we consider is given in \cite{fukai1997simple}. Note that unlike the common description of WTA dynamics as a system of competing neurons, we here have a system of `competing' kernel Fourier coefficients. The overall dynamics are as follows: if we initialize the kernel with small initial values, its vec-2D Fourier coefficients $|(\bs{Q k})_i|$ will also be small. For each input mode $\alpha$, we have a number of associated frequency indices $j$, and the corresponding terms $|(\bs{Q k})_j|$ each contribute to the respective effective singular value $a_{\alpha}$ (Eq. \ref{eq:a_sum_of_freq_main}). Initially,  $a_{\alpha}$ is much smaller than $s_{\alpha}$. Therefore, the values $|(\bs{Q k})_j|^2$ initially grow exponentially (Eq. \ref{eq:diag_winner_takes_all_main} when $a_{\alpha} \approx 0$). However, they do so with \emph{different} exponents; within the set of frequencies associated to the mode $\alpha$, the difference lies in the factors $ |(\bs{Q} \bs{\phi}^{\alpha})_j|$. As soon as the values $|(\bs{Q k})_j|^2$ start growing, the self-inhibition as well as the mutual inhibition between the coefficients associated to the same mode starts to take off. In practice, the coefficients $|(\bs{Q k})_j|$ with the largest factors $|(\bs{Q} \bs{\phi}^{\alpha})_j|$ very strongly inhibit the growth of the other coefficients, such that only the former coefficients grow and significantly contribute to the effective singular value $a_{\alpha}$. Eventually, $a_{\alpha}$ saturates to its final value  $s_{\alpha}$, and all derivatives become zero. Depending on the factors  $ |(\bs{Q} \bs{\phi}^{\alpha})_j|$, we thus end up with one or more coefficients $|(\bs{Q k})_j|$ that have `won' the competition; this means the final kernel consists of a sum of those frequencies only. \\
While in a linear, fully connected network, the weight matrices take on the inherent structure of the dataset (see Sec.~\ref{sec:compare_FCNN}); the kernel in our linear CNN only picks up the most dominant frequencies associated to each mode, i.e., singular vector associated to the dataset. It thus acts as a filter; not necessarily a low-pass filter, but a filter of the most dominant frequencies. The results of an experiment with two classes, each a sum of pure cosines with different amplitudes, is illustrated in Fig. \ref{fig:CNN_dominant_freq}. \\
For more general datasets, the singular vectors can in general share the same frequency. Formally, $|\bs{Q} \bs{\phi}^{\alpha}|_j$ can be significant for different modes $\alpha$. In this case, Eq. \ref{eq:diag_winner_takes_all_main} does not hold and we have to revert back to the more general equation Eq. \ref{eq:update_diag_Qk_transformed}. The experiments discussed in the next section show that in those more general cases a form of the dominant frequency filtering is still present. However, the dominancy of a frequency is now determined through another factor as well: assume two values  $|\bs{Q} \bs{\phi}^{\beta}|_j$ and $|\bs{Q} \bs{\phi}^{\gamma}|_j$, with $\beta < \gamma$, are significant. During the discovery of mode $\beta$ the kernel Fourier coefficient $|(\bs{Q k})_j|$ develops. Later, at the start of the discovery of the mode $\gamma$, the coefficient $|(\bs{Q k})_j|$ starts from a higher value than coefficients $|(\bs{Q k})_{j'}|$ that were not developed before, and that still have their small, random initial value. Loosely speaking, this gives the coefficient $|(\bs{Q k})_j|$ a `competitive advantage'. This also means that frequencies that are dominant in the first few modes (thus high $|\bs{Q} \bs{\phi}^{\alpha}|_j$ for the first modes $\alpha$) are the frequencies that are more likely to be important for the final kernel. 

\section{Experiments with More General Datasets}\label{sec:experiments}
\begin{figure*}[h!]
\centering
\includegraphics[width=\textwidth]{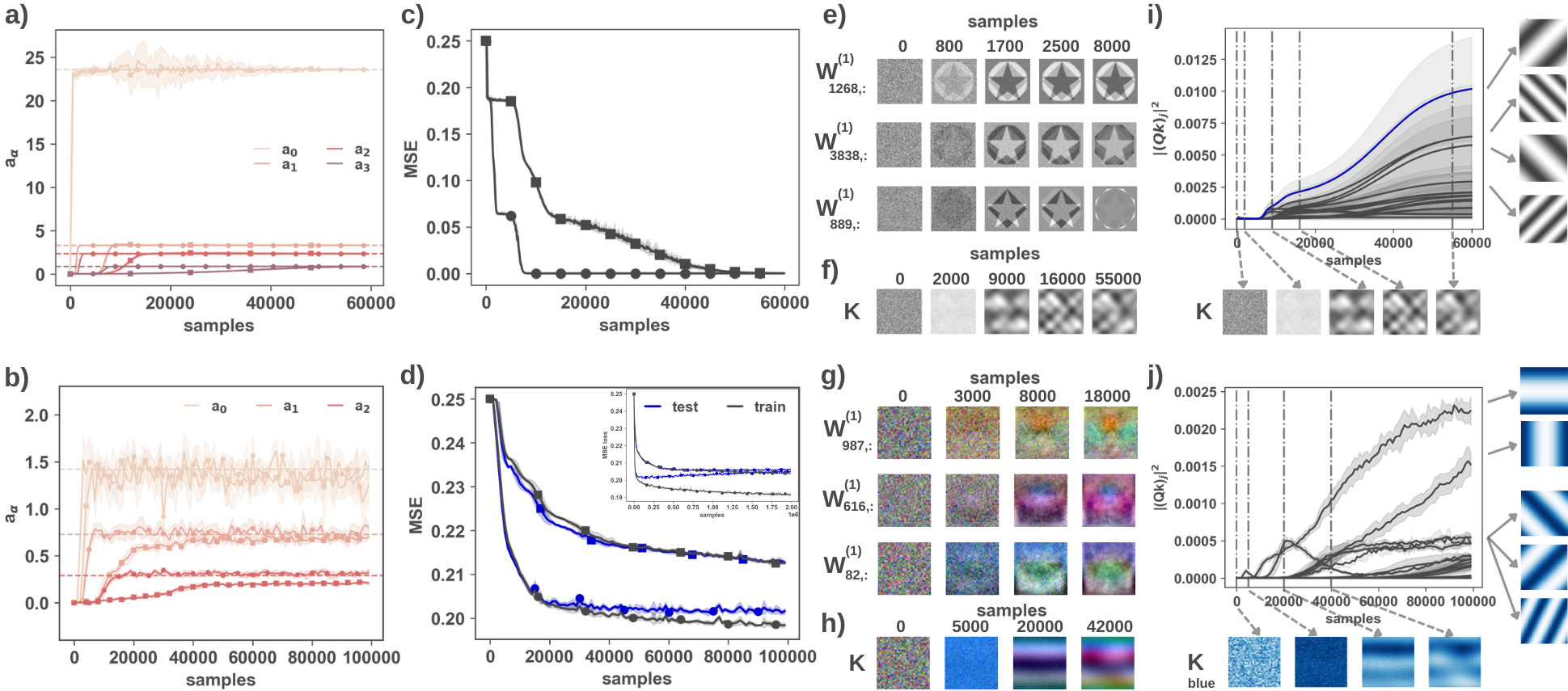}
\caption{ Top row: geometric shapes. Bottom row: CIFAR-4. $\bs{a, b)}$ Evolution of the effective singular values for the CNN (square markers) and the FCNN (round markers). $\bs{c, d)}$: MSE loss for the CNN (square markers) and the FCNN (round markers). For CIFAR-4, the test loss is plotted in blue. The inset in d) shows the loss up to $10^6$ timesteps. $\bs{e, g)}$ Evolution of the weights connected to 3 randomly selected nodes in the hidden layer of the FCNN, reshaped as $n \times n$ matrices. Shown timepoints (in number of samples) roughly correspond to the moments each mode is fully discovered (compare to subfigures a) and b) ). $\bs{f,h)}$ Evolution of the kernel at analogous timepoints for the CNN. $\bs{i,j)}$ Evolution of the coefficients $|(\bs{Q k})_j|^2$ (middle), frequencies corresponding to highest $|(\bs{Q k})_j|^2$ of final kernel (right). Only the kernel of the blue channel is shown for CIFAR-4. $\bs{a-f)}$ Solid lines: averages over experimental runs. Shaded regions: standard deviations. Vertical dashed lines: selected timepoints. Horizontal dashed lines: singular values $s_\alpha$. The kernels and partial weight matrices shown, are randomly selected across experiment runs.}
\label{fig:results_experiments}
\end{figure*}

We now report the results of training a linear CNN from small, random initial conditions on two more general datasets: the dataset of geometric shapes (see Fig. \ref{fig:geometric_dataset_structure}) and on the dataset consisting of the first 4 classes of CIFAR-10 (`airplane', `automobile', `bird' \& `cat'), which we will call CIFAR-4. We subtract the first mode, i.e. the average over images, from the CIFAR-4 dataset.  Therefore only the 3 modes needed to distinguish between 4 classes remain (SI Fig. \ref{fig:modes_cifar_4}). CIFAR-4 has multiple samples per class: $\bs{\Sigma^{xx}}$ is no longer diagonal in the basis given by $\bs{V}$ ( Eq. \ref{eq:sigma_x_x_V} does not hold), which influences the dynamics through a coupling of the modes. Moreover, we can hold out a separate test set to track the test loss for CIFAR-4. We train the CNN with learning rate $\lambda$, and the FCNN with a learning rate $\lambda_{FC} = n \lambda$. The results are shown in Fig. \ref{fig:results_experiments}; experimental details see \hyperref[sec_sup:experiments]{SI}. \\ 
We can conclude that the general insights we derived before are still valid. First of all, in both linear CNNs and linear FCNNs the modes of $\bs{ \Sigma^{ y x}}$ are discovered with rapid, successive transitions (Fig. \ref{fig:results_experiments} (a) and (b)). However, the CNN exhibits a different effective learning rate for each mode (see Sec. \ref{sec:learning_dynamics}). Since the first mode of the geometric shape dataset is essentially the average over the shapes (see Fig. \ref{fig:geometric_dataset_structure}), and an average corresponds to the zero-frequency, the CNN does not show an additional delay for this mode ($d_0 \approx 1$). All other mode discoveries have an additional delay with respect to the trajectories for the FCNN.
Fig. \ref{fig:results_experiments} (c) and (d) show the evolution of the loss. The loss for CIFAR-4 does not go to zero, meaning there are samples that cannot be classified by a linear CNN or FCNN. The FCNN has discovered all the modes after around 18000 samples, and at that point reaches a train and test loss of around 0.21. After that, it starts to overfit. The CNN needs a lot more samples to discover all the modes, but doesn't start to overfit. The latter might be due to the implicit regularisation given by the dominant frequency bias: Fig. \ref{fig:results_experiments} (e) and (g) show that the rows of the first weight matrix in FCNN are mixtures of the discovered singular vectors, as given by Eq. \ref{eq:W1_bar}. Fig. \ref{fig:results_experiments} (f) and (h), on the other hand, show that, at any point during training, the kernel is a sparse mixture of the dominant frequencies of the singular vectors discovered so far.  Fig. \ref{fig:results_experiments} (i) and (j) show the evolution of the kernel Fourier coefficients $|(\bs{Q k})_j|$ in detail. In the plot for the geometric shapes dataset, we have highlighted a trajectory in blue. This cascaded trajectory is an example of how a frequency initially becomes dominant through a high value $|\bs{Q} \bs{\phi}^{\alpha}|_j$ during the discovery of mode $\alpha$, and subsequently can remain dominant due to its higher value at the start of the discovery of a subsequent mode. That the kernel frequencies are really the dominant frequencies is further illustrated in the \hyperref[sec_sup:compare_spectra]{SI}.

\section{Experiments with Deep, Non-Linear CNNs}\label{sec:nonlinear}

\begin{figure*}[h!]
\centering
\includegraphics[width=\textwidth]{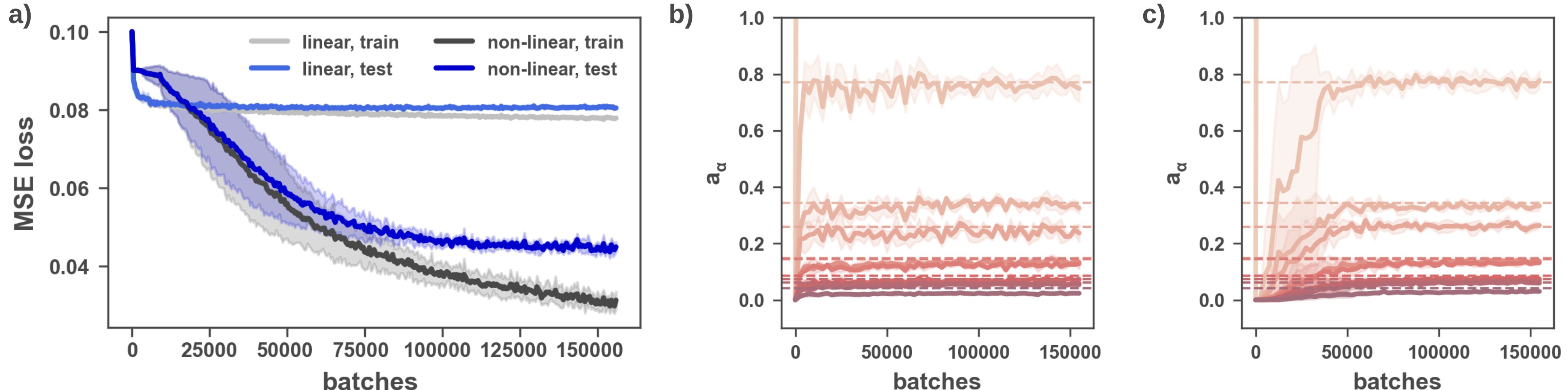}
\caption{Results of experiments with deep, linear and non-linear CNNs trained on CIFAR-10. a) Evolution of the train and test MSE loss. b) Evolution of the effective singular values for the linear CNN. c) Evolution of the effective singular values for the non-linear CNN. a-c) Solid lines: average over 5 trials from random initial conditions; shaded regions: standard deviations. Horizontal dashed lines: singular values $s_\alpha$. In fig. b) and c) we limit the y-axis to show more detail of the smaller effective singular values starting from $a_1$.}% The value of $a_0$ is much larger and in both cases quickly saturates at the value 8.34 (not shown).}
\label{fig:loss_nonlinear}
\end{figure*}

% Finally, we now briefly report the results of experiments with deeper, linear and non-linear CNNs trained on CIFAR-10. While our theory does not cover depth and non-linearity, the experiments indicate that these networks still discover the statistical dataset structure in an ordered manner, from broader to finer distinctions. \\
We now consider deep linear and non-linear CNNs trained on CIFAR-10; both have the same overall architecture, but the non-linear network has ReLU activation functions and additional max-pooling layers. The overall network architecture consists of four convolutional layers, each with 16 channels; a flatten layer, and two fully-connected layers. The kernel size is $8\times 8$, while the image size is $32 \times 32$: this means these models not only exhibit weight sharing but they also exhibit locality. The additional max-pooling layers will make the non-linear network (partially) translation invariant. The updates now take place in batches, we use multiple channels per layer, and we use zero-padding for the convolutions (details see \hyperref[sec_sup:deep_networks]{SI}). In fig. \ref{fig:loss_nonlinear}, we show the results of the experiments. Fig. \ref{fig:loss_nonlinear} (a) shows the evolution of the loss for the linear and non-linear model, averaged over trials. Fig.  \ref{fig:loss_nonlinear} (b) and (c) show the corresponding evolutions of the effective singular values for the linear and non-linear CNN, respectively. Firstly, from fig. \ref{fig:loss_nonlinear} (b) and (c) we can see that both networks discover the statistical dataset structure with ordered, stage-like transitions. However, the evolution of the effective singular values for the  non-linear network are delayed with respect to the linear network. Secondly, the non-linear CNN exhibits a lower loss, and given that both types of networks eventually exhibit the same, saturated effective singular values, this lower loss cannot be explained from the discovery of the statistical structure of the dataset alone. We hypothesize that while the non-linear model discovers the statistical structure, it at the same time discovers different aspects of the dataset structure as well. These aspects are likely not captured by an average over samples (Eq. \ref{eq:sigma_y_hat_x}), which could explain why they are not captured by the effective singular values $\bs{a_\alpha}$ (Eq. \ref{eq:SVD_sigma_y_hat_x}) alone. 

%towards the ground truth singular values $\bs{s_\alpha}$: recall that the matrix  $\bs{\Sigma^{y  x}} = \bs{U} \bs{S} \bs{V}^T$ on each row contains the averaged class image, i.e., the class mean. In general, the samples of each class are distributed around this class mean, and these distributions can have complex structures. We conjecture  the non-linearities in the network leverage some of this more complex structure. But this cannot be captured by the evolution of the effective singular values, as these are related to the class means only.  

% The first additional max-pooling layer in the non-linear network is placed after the first two convolutional layers, and the other one is placed after the next two convolutional layers. We furthermore use zero-padding at the boundaries for the convolutions, which is more conventional that the wrapping of the image we assumed in the theoretical derivations. Finally, to stay closer to previous, theoretical results, we again use the MSE loss and we keep the last layer linear, even in the non-linear model.

\section{Discussion}
The above results shed new light on the use of convolutions in neural networks. Our theoretical analysis uses a linear CNN and a number of assumptions (listed in Sec. \ref{sec:assumptions}).  We first discussed how the relevant structure of the dataset is captured by the singular vectors of the input-output correlation matrix, which generally represent broader to finer distinctions between classes. We then showed how the dynamics of learning are influenced by the interplay between the singular vectors associated to the dataset structure and the singular vectors associated to the convolutional network structure. We subsequently showed that linear CNNs discover the statistical dataset structure with rapid, stage-like transitions, and they do so at an effective learning rate that depends on this interplay. Moreover, this interplay yields a dominant frequency bias: internally, only the most dominant frequencies of each singular vector of the dataset are used by the CNN to discern between classes. It is thus both the statistical structure of the dataset, embodied by the corresponding singular vectors, as the convolutional structure of the network---for which the singular vectors are frequencies---that influence the learning. Our experiments show that these conclusions broadly hold when using more general datasets. We moreover show that deep, non-linear CNNs also discover the statistical structure of the dataset in an ordered, stage-like fashion (see also \citealt{hacohen2022principal}). However, they at the same time seem to discover aspects of the dataset structure that are not captured by the statistical structure.\\
% We would need an extension of our theory of learning that incorporates depth, locality, max-pooling and/or non-linear activation functions to analyze more precisely what happens in these deep, non-linear networks internally. E.g., 
In our results, the statistical structure of the dataset is discovered through \emph{time}. Deeper, non-linear CNNs are known to process higher-order visual relationships in later layers as well: i.e., there is an additional notion of hierarchy through depth \citep{krizhevsky2012imagenet, simonyan2013deep, olah2017feature}. An extension of our theory could clarify how this hierarchy over depth develops over time, and how this relates to the dominant frequency bias and results on the relationship between learning in linear and non-linear CNNs \citep{kalimeris2019sgd, refinetti2022neural}.
The dominant frequency bias we find is a form of implicit regularization that could help explain why large CNNs sometimes generalize well, even without additional regularization. It is seemingly at odds with the `spectral bias', which states that lower frequencies are learned first \cite{xu2018understanding, rahaman2019spectral, xu2019training, ronen2019convergence, basri2020frequency,cao2021towards}.
We claim on the other hand that the bias is towards frequencies that are dominant in the Fourier spectrum of the singular vectors, irrespective of whether these are actually high or low frequencies. One thing to note is that broader distinctions in shape correspond to lower frequencies. We find that broader distinctions are uncovered first, and that therefore lower frequencies are learned first. How the two concepts exactly relate is the topic of future work. Finally, our results could also help explain why CNNs are sensitive to frequency domain perturbations \citep{jo2017measuring, tsuzuku2019structural}, and why they often rely on textures (images with a sparse frequency spectrum) instead of shapes \citep{baker2018deep, geirhos2018imagenettrained,BrendelB19}.

%, and why they have the tendency to learn surface statistical regularities as opposed to higher level abstractions \citep{jo2017measuring}.\\

%Another interesting empirical observation is that CNN seem to rely much more on texture than on global shape to classify images \citep{geirhos2018imagenettrained, baker2018deep, BrendelB19}. This idea is illustrated in fig.  \ref{fig:texture_bias}. In another but related line of work, the authors show that CNN seem to use what they call ‘surface statistical regularities' instead of shape \citep{jo2017measuring}. The authors construct different versions of the same image dataset; while for each version, the (human) recognizability of the objects is almost entirely preserved, they manipulate the Fourier spectrum of each image through a different, fixed mask (see fig. \ref{fig:surface_regularities}). They find that CNN trained on one of the versions of the dataset, does not generalize well to other versions of the data (i.e., with a different Fourier mask, or different ‘surface statistical regularities) despite the high level concepts remaining the same. 

\section*{Acknowledegments}
H.P. and J.L. acknowledge fellowships from the Research Foundation Flanders under Grant No.11A6819N and 11G1621N. V.G. acknowledges support from Research Foundation Flanders under Grant No.G032822N and G0K9322N.

%This is another form of hierarchy in the dataset that cannot be described by the statistical structure, i.e.. the SVD of the input-output correlation matrix, and is therefore not captured by our theory. However, it is interesting to note that the reported shape of the kernels in the early layers of deep, non-linear networks (\textcolor{red}{ref}) resemble the shape we would expect from linear CNNs, namely sparse frequency patterns (see \textcolor{red}{ref fig}). An extension of our theory could clarify how this more intricate hierarchy over depth then develops over time.\\

\clearpage

\bibliography{sample-base}
\bibliographystyle{icml2023}

%%%%%%%%%%%%%%%%%%%%%%%%%%%%%%%%%%%%%%%%%%%%%%%%%%%%%%%%%%%%%%%%%%%%%%%%%%%%%%%
%%%%%%%%%%%%%%%%%%%%%%%%%%%%%%%%%%%%%%%%%%%%%%%%%%%%%%%%%%%%%%%%%%%%%%%%%%%%%%%
% APPENDIX
%%%%%%%%%%%%%%%%%%%%%%%%%%%%%%%%%%%%%%%%%%%%%%%%%%%%%%%%%%%%%%%%%%%%%%%%%%%%%%%
%%%%%%%%%%%%%%%%%%%%%%%%%%%%%%%%%%%%%%%%%%%%%%%%%%%%%%%%%%%%%%%%%%%%%%%%%%%%%%%
\newpage
\appendix
\onecolumn
\renewcommand{\theequation}{S.\arabic{equation}}
\setcounter{equation}{0}
\renewcommand{\thefigure}{S.\arabic{figure}}
\setcounter{figure}{0}

\section{Diagonality of $\bs{\Sigma^{ x  x}}$ in the Basis Given by $\bs{V}$ and Relationship to Class Coherence} \label{sec_sup:sigma_x_x}

 We here discuss the diagonality of $\bs{\Sigma^{ x  x}}$ in the basis given by $\bs{V}$, and its relationship to class coherence. We start from the assumption that the label vectors are orthonormal, $\bs{y}_{c_p}^T \bs{y}_{c_q} = \delta_{c_pc_q}$, where $c_p$ and $c_q$ are class indices. This is the case when the classes are one-hot encoded. Then, a property of the singular value decomposition of a general, real matrix $\bs B$ is that the right singular vectors of this matrix are the eigenvectors of the matrix $\bs{B}^{T}\bs{B}$. We can make use of the orthonormality of the label vectors and general properties of the kronecker product (denoted $\otimes$) to derive:
\begin{align}
    \bs{\Sigma^{y  x}}^{T}  \bs{\Sigma^{y  x}} &=  \langle  \bs{y} \otimes \bs {x}^T \rangle ^{T} \langle  \bs{y} \otimes \bs {x}^T \rangle \\
        &=  \langle   \bs {x} \otimes \bs{y}^T \rangle  \langle  \bs{y} \otimes \bs {x}^T \rangle \\
        &= \frac{C^2}{N^2} \sum_{c_r=0}^{m-1} \sum_{c_q=0}^{m-1} ( \langle \bs x \rangle_{c_r} \otimes y_{c_r} ^T )  ( y_{c_q} \otimes  \langle \bs x^T \rangle_{c_q}  ) \\
        &= \frac{C^2}{N^2} \sum_{c_r=0}^{m-1} \sum_{c_q=0}^{m-1} y_{c_r} ^T y_{c_q} ( \langle \bs x \rangle_{c_r} \otimes  \langle \bs x^T \rangle_{c_q}  ) \\
        &= \frac{C^2}{N^2} \sum_{c_r=0}^{m-1}  \langle \bs x \rangle_{c_r} \otimes  \langle \bs x^T \rangle_{c_r}  
        \label{eq:phi}
\end{align}
where $ \langle \rangle_{c_r}$ denotes the average over all samples belonging to the same class ${c_r}$, $p$ is the number of classes, $N$ is the total number of samples, and $C$ is the number of samples per class (we assume a balanced dataset). If we define $\bs{\Phi}^{xx} =  \frac{C^2}{N^2} \sum_{c_r=0}^{m-1}  \langle \bs x \rangle_{c_r} \otimes  \langle \bs x \rangle_{c_r}  $, we thus have that $\bs{V}^{T}  \bs{\Phi}^{xx}  \bs{V} $ is diagonal: the right singular vectors of $\bs{\Sigma^{y  x}} $ are the eigenvectors of $\bs{\Phi}^{xx} =  \bs{\Sigma^{y  x}}^{T}  \bs{\Sigma^{y  x}} $.  The corresponding eigenvalues are equal to the square of the singular values, thus the diagonal values of  $\bs S^T \bs S$. \\

$\bs{\Sigma^{ x  x}} = \langle  \bs{x} \otimes \bs {x}^T \rangle $  and $\bs{\Phi}^{xx} $  (eq. \ref{eq:phi}) are different in general, but they are close to equal if samples within a class are very similar (at least in pixel-space):
\begin{align}
    \bs x^{(i)} = \langle \bs x \rangle_{c_r} \, +  \bs{\epsilon}^{(i)}, \;  \;  \; \lVert \bs{\epsilon}^{(i)} \rVert << \lVert \langle \bs x \rangle_{c_r} \rVert
    \label{eq:x_similar_to_avg}
\end{align} 
where $\bs x^{(i)}$ is the sample with index $i$ belonging to a class with index $c_r$. We have:
\begin{align}
    \bs{\Sigma^{ x  x}} &= \langle  \bs{x} \otimes \bs {x}^T \rangle \\
    &= \big \langle  \,( \langle \bs x \rangle_{c_r} \, +  \bs{\epsilon}^{(i)}) \otimes  \,( \langle \bs x \rangle_{c_r} \, +  \bs{\epsilon}^{(i)})^T \big \rangle\\
    &= \big \langle  \,( \langle \bs x \rangle_{c_r} \otimes   \langle \bs x \rangle_{c_r}^T) + ( \langle \bs x \rangle_{c_r} \otimes  \bs{\epsilon}^{(i)}) +... \big \rangle
\end{align}
Thus, if $\bs x^{(i)} \approx \langle \bs x \rangle_{c_r}$ (eq. \ref{eq:x_similar_to_avg}), then:
\begin{align}
    \bs{\Sigma^{ x  x}} &\approx \big \langle  \,( \langle \bs x \rangle_{c_r} \otimes   \langle \bs x \rangle_{c_r}^T) \big \rangle \\
    &= \frac{C}{N} \sum_{c_p=0}^{m-1}  \langle \bs x \rangle_{c_r} \otimes  \langle \bs x \rangle_{c_r}^T   \\
    &= \frac{N}{C} \bs{\Phi}^{xx} \\
    &= p \bs{\Phi}^{xx}.
\end{align}
In other words, if samples within a class are very similar (=high class coherence), we can use the same eigen-/right singular vectors to decompose $\bs{\Sigma^{ x  x}}$ and  $\bs{\Sigma^{ y  x}}$ into a sum of their respective modes. Thus the higher the class coherence, the more the two matrices represent the same inherent structure. If $\bs{\Sigma^{ x  x}}$ is exactly equal to $p \bs{\Phi}^{xx} $, the $p$ eigenvalues of $\bs{\Sigma^{ x  x}}$ are given by the $p$ non-zero diagonal elements of $p \bs S^T \bs S$.

\section{Definition of Q} \label{sec_sup:Q}

We first recall the definition of the \emph{one}-dimensional discrete Fourier transform (DFT) matrix, denoted by the $n \times n$ matrix $\bs{F}^{(n)}$:
\begin{equation}
(\bs{F}^{(n)})_{p,q} = (\omega_n)^{pq}, \; \omega_n = \exp(-2 \pi i / n),
\end{equation}
where the superscript $pq$ denotes the product of indices $p$ and $q$ applied as an exponent, and $i^2 = -1$. Then the $n^2 \times n^2$ complex-valued matrix $\bs{Q}$ is given by:
\begin{equation}
 \label{eq:Q}
   \bs{Q} = \frac{1}{n} \bs{F}^{(n)} \otimes \bs{F}^{(n)}
\end{equation}
with $\otimes$ the kronecker product. $\bs{Q}$ is unitary ($\bs{Q^{*T}} = \bs{Q^{-1}}$) and symmetric.

\section{Doubly Block Circulant Matrices} \label{sec_sup:dbc}

\subsection{Doubly Block Circulant Matrix Definition}
$circ(\bs b)$: circulant matrix of the vector $\bs b$ with length $q$: 

\begin{equation}
circ([b_0, b_1, \cdots, b_{q-2}, b_{q-1}]^T) = 
\begin{bmatrix}
b_0 & {b_{q-1}} & \cdots & b_{2} & b_{1} \\
{b_{1}} & b_0 & b_{q-1} & \cdots  & b_{2}\\
{b_{2}} & {b_{1}} & b_0 & b_{q-1} & \cdots  \\
\vdots &  &  & \ddots  &  \\
b_{q-1} &  & \dots &   &  b_0 \\
\end{bmatrix}
\end{equation}

then $dbc(\bs{B})$, the doubly block circulant matrix of the matrix $\bs B$, is given by  \cite{sedghi2019singular}:

\begin{equation}
dbc(\bs{B}) = \begin{bmatrix}
circ(B_{0,:}) & circ(B_{u-1,:}) & \dots &circ(B_{1,:})\\
circ(B_{1,:}) & circ(B_{0,:}) & \dots &circ(B_{2,:})\\
\vdots &  & \ddots &   &  \\
circ(B_{u-1,:}) &  & \dots & circ(B_{0,:})\\
\end{bmatrix}
\label{eq:dbcirc}
\end{equation}
where $u$ is the number of rows of $\bs B$, and $\bs B_{i,:}$ is the $i^{th}$ row. Note that this is the definition for an actual convolution, see below. 

\subsection{Convolution and correlation: notes on flipping the kernel or the image}

Given a kernel $\bs{K}$ and an image $\bs{X}$, the convolution operation is defined as: 
\begin{align}
    &(\bs{X} \ast \bs{K})_{i,j} = \sum_{m} \sum_{l} \bs{X}_{m,l} K_{i-m, j-l} 
\end{align}
or, equivalently (since convolution is commutative):
\begin{align}
    &(\bs{K} \ast \bs{X} )_{i,j} = \sum_{m} \sum_{l} \bs{X}_{i-m, j-l} K_{m,l}.
\end{align}
This implies the kernel (or the image) is applied in a `flipped' manner, i.e., when we increase $m$ and $l$, we decrease the corresponding indices. This amounts to flipping the kernel (or image) over the x- and y-axis, or equivalently, rotating it 180 degrees, before applying the convolution. \\
In practice, however, convolutional layers are usually implemented as correlations (here denoted $\star$):
\begin{equation}
    (\bs{X} \star \bs{K})_{i,j} = \sum_{m} \sum_{l} \bs{X}_{m,l} K_{i+m, j+l}
\end{equation}
thus without flipping the kernel beforehand. This makes the operation more intuitive (but it's not commutative). Whether the kernel is flipped beforehand or not, the learning algorithm will learn
the appropriate values of the kernel in the appropriate place: the result of learning with correlations instead of convolutions is just a flipped kernel \cite{Goodfellow-et-al-2016}. Most literature therefore uses the term ``convolution" in lieu of ``correlation". The actual operation used influences the definition of the doubly block circulant matrices and the gradient descent equations, however, and so it's important to keep track of the details when comparing different sources.   \\
As it turns out, we can define a very similar circulant and doubly block circulant matrix to replace a (circular) correlation (see the definitions used in \cite{sedghi2019singular}):
\begin{equation}
    circ_{corr}([b_0, b_1, \cdots, b_{q-2}, b_{q-1}]^T) = 
    \begin{bmatrix}
    b_0 & {b_{1}} & \cdots & b_{q-2} & b_{q-1} \\
    b_{q-1} & b_0 & b_{1} & \cdots  & b_{q-2}\\
    {b_{q-2}} & {b_{q-1}} & b_0 & b_{1} & \cdots  \\
    \vdots &  &  & \ddots  &  \\
    b_{1} &  & \dots &   &  b_0 \\
    \end{bmatrix}
    \label{eq:circ_corr}
\end{equation}
and 
\begin{equation}
dbc_{corr}(\bs{B}) = \begin{bmatrix}
circ(B_{0,:}) & circ(B_{1,:}) & \dots &circ(B_{u-1,:})\\
circ(B_{u-1,:}) & circ(B_{0,:}) & \dots &circ(B_{u-2,:})\\
\vdots &  & \ddots &   &  \\
circ(B_{1,:}) &  & \dots & circ(B_{0,:})\\.
\end{bmatrix}
\label{eq:dbcirc_corr}
\end{equation}
When we compare those definitions to the earlier defined matrices for circular convolutions, we see that the difference lies in a permutation of the indices equal to a reversion and a shift, e.g., for a (column) vector $\bs{b}$:
\begin{align}
    rev(\bs{b}^T) = [ b_{q-1}, b_{q-2}, \cdots, b_{0} ] 
\end{align}
and 
\begin{align}
    shift_{\rightarrow 1} (rev(\bs{b}^T)) = [ b_{0}, b_{q-1}, \cdots, b_{1} ],
\end{align}
yielding
\begin{equation}
    circ_{corr}(\bs{b} ) = circ(shift_{\rightarrow 1} (rev(\bs{b}))),
\end{equation}
and a similar permutation of the rows for the doubly block circulant matrices. Due to the properties of Fourier transforms, we can also flip the kernel through applying the Fourier transform twice:
\begin{equation}
    \bs{Q} \bs{Q} \bs{k} =  shift_{\rightarrow 1} (rev(\bs{k}^T))
    \label{eq:fourier_flip}
\end{equation} 
This allows us to translate all the derivations from implementations with convolutions to implementations with correlations: where 
\begin{equation}
    \bs{dbc}(\bs{K}) =  n \, \bs{Q}^{-1}  diag(\bs{Q k} ) \bs{Q} =  n \, \bs{Q}  diag(\bs{Q^{-1} k} ) \bs{Q}^{-1}
\end{equation}
(since the dbc matrix is real valued, thus $\bs{dbc}(\bs{K})  = \bs{dbc}(\bs{K})^*$ and $\bs{Q}^{-1}  = \bs{Q}^*$), we can use eq. \ref{eq:fourier_flip} to write:
\begin{equation}
    \bs{dbc}_{corr}(\bs{K}) =  \bs{dbc}(\bs{K}_{flip}) =  n \, \bs{Q}  diag(\bs{Q^{-1} (QQk)} ) \bs{Q}^{-1} 
\end{equation}
thus
\begin{equation}
     \bs{dbc}_{corr}(\bs{K}) = \bs{dbc}(\bs{K}_{flip}) =  n \, \bs{Q}  diag(\bs{Qk} ) \bs{Q}^{-1}.
\end{equation}

% ons, padding, flipped kernels, kernel size etc, see also \cite{sedghi2019singular} "The quality of
% this approximation has been heavily analyzed in the case of one-dimensional signals (Gray, 2006)."; section "6.2. Circulant Projection vs. 2D Convolution" in  \cite{cheng2015exploration}; flipped kernels in "deep learning book", chapter convolutional neural networks]

\section{Derivation of the Differential Equations of Gradient Descent } \label{sec_sup:diff_eq}

\subsection{Definition of the network and learning algorithm}

As described in the main paper, we consider a convolutional network where the predictions $\bs{\hat y}$ are given by:
\begin{align}
   \bs{ \hat y} & =  \bs W \bs{dbc(k)} \bs{x} 
\end{align}
here $\bs{x} = vec(\bs{X})$ is the vectorized input image, $\bs{k} = vec(\bs{K})$ is the vectorized kernel, and $\bs{W}$ is the weight matrix of the fully connected layer. \\
The loss is given by the MSE loss, i.e.,  
\begin{equation}
    L = \frac{1}{2} \sum_{l=0}^p (y_l - \hat y_l)^2.
\end{equation}
The kernel and the weights are updated according to the gradients of this loss:
\begin{align}
\label{eq_sup:delta_k}
    & \Delta \bs k =  - \lambda  \frac{\partial L }{\partial \bs k^T} 
    \\
   & \Delta \bs W = - \lambda  \frac{\partial L }{\partial \bs W^T}
   \label{eq_sup:delta_w}
\end{align}
where $\lambda$ is the learning rate. 

\subsection{Gradient descent for the convolutional layer}

For the convolutional layer, the gradient of the loss with respect to the kernel is in itself given by a convolution, but with a flipped image (or, equivalently, an image rotated 180 degrees). We can write this equation using doubly block circulant matrices: 
\begin{align}
     \frac{\partial L }{\partial \bs k^T} 
     &= vec \big ( \bs{X}_{flip} \circledast \frac{\partial L }{\partial \bs H^T} \big ) \\
    &= dbc(\bs{X}_{flip}) \frac{\partial L }{\partial \bs h^T},
\end{align}
with
\begin{align}
    \frac{\partial L }{\partial \bs h^T} = - \sum_{l=0}^p (y_l - \hat y_l)  (\bs W^T)_{:,l}.
\end{align}
Here $\bs{H}$ denotes the activity of the hidden layer, and $\bs{h} = vec(\bs{H})$. $(\bs{W}^T)_{:,l}$ denotes column $l$ of $\bs{W}^T$.

We will now rewrite this equation in a form that will turn out to be useful in the next steps: 
\begin{align}
   &  \frac{\partial L }{\partial \bs k^T}  =  n \, \sum_{l=0}^p   \bs{Q}   diag(\bs{Q x} ) \bs{Q}^{-1} (y_l - \hat y_l)  (\bs W^T)_{:,l} \\
    &\iff  \big (\frac{1}{n}  \bs Q^{-1} \frac{\partial L }{\partial \bs k^T} \big )_j =   \sum_{l=0}^p    diag(\bs{Q x} )_{j,:} (y_l - \hat y_l)  (\bs {Q^{-1} W}^T)_{:,l} \\
     &\iff \big (\frac{1}{n}  \bs Q^{-1}  \frac{\partial L }{\partial \bs k^T} \big )_j =  \sum_{l=0}^p    (\bs{Q x})_j (y_l - \hat y_l)  (\bs {Q^{-1}  W}^T)_{j,l} \\
    &\iff \big (\frac{1}{n}  \bs Q^{-1}  \frac{\partial L }{\partial \bs k^T} \big )_j =  \sum_{l=0}^p  \big( (\bs y - \bs{\hat y}) \otimes (\bs{Q x})^T \big)_{l,j} (\bs {Q^{-1}  W}^T)_{j,l} \\
    &\iff \frac{1}{n}   \big (\bs Q^{-1}  \frac{\partial L }{\partial \bs k^T} \big )_j =  (\bs {Q^{-1}  W}^T)_{j,:} \big( (\bs y - \bs{\hat y}) \otimes \bs{x}^T\big) \bs{Q}^T _{:,j} 
    \label{sup_eq:grad_k}
\end{align}

\subsection{Gradient descent for the fully connected layer}
For the fully connected layer with weights $\bs W$, we can write down the gradient in the conventional form:
\begin{align}
      \big(\frac{\partial L }{\partial \bs W^T}\big)_{l,:}  &=  (y_l - \hat y_l) \bs{x}^T dbc(\bs k)^T 
\end{align}
and, for future use, rewrite this as:
\begin{align}
\big( \frac{\partial L }{\partial \bs W^T}\big )_{l,:} 
    & =   (y_l - \hat{y}_l) \bs{x}^T n \bs{Q}^T diag(\bs{ Q k})^T \bs{Q}^{-T} \\
      \iff  \frac{1}{n} \big( \frac{\partial L }{\partial \bs W^T}  \big)_{l,:}  & =  \big( (\bs y - \bs{\hat y}) \otimes \bs{x}^{T}  \big)_{l,:} \bs{Q} diag(\bs{ Q k}) \bs{Q}^{-1}
        \label{sup_eq:grad_w}
\end{align}
where we used the fact $\bs{Q}^{-1} = \bs{Q}^{-T}$ and $\bs{Q} = \bs{Q}^T$.

\subsection{From discrete gradient descent to differential equations}

The equations \ref{sup_eq:grad_k} and \ref{sup_eq:grad_w}, together with the update rules given by eq.  \ref{eq_sup:delta_k}   and \ref{eq_sup:delta_w}, tell us how to update the kernel and the weight matrix at each discrete time step $\Delta t$. We will proceed to make a continuous approximation of these discrete updates. This is a valid approximation as long as the learning rate is small enough. In this slow learning regime, we can assume the weights only change minimally over a number of updates with different samples. The factors $\big( (\bs y - \bs{\hat y}) \otimes \bs{x}^T\big)$ can then be replaced by their average over those samples (see \cite{saxe2019mathematical}). In the main body, we  defined the matrices $\bs{\Sigma^{ x  x}} = \langle  \bs{x} \otimes \bs {x}^T \rangle$ and $\bs{\Sigma^{ y  x}} = \langle  \bs{y} \otimes \bs {x}^T \rangle$, which can be computed from the dataset alone. We also defined $ \bs{\Sigma^{\hat y  x}}  = \langle \bs{ \hat y \otimes x^T } \rangle $. These definition allow us to rewrite the difference eqs. \ref{sup_eq:grad_k}, \ref{sup_eq:grad_w},  \ref{eq_sup:delta_k}   and \ref{eq_sup:delta_w},  as:
\begin{align}
   & \frac{1}{n \lambda}  \big ( \frac{d \, diag(\bs{Q^{-1} k})}{dt}  \big )_{j,j} = \bs Q^{-1}_{j,:} \bs{W}^T\big(  \bs{\Sigma^{ y  x}} - \bs{\Sigma^{ \hat{y}  x}}    )  \bs Q_{:,j}  \\
    &    \frac{1}{n\lambda} \big( \frac{d \bs W }{dt}  \big)  =  \big(  \bs{\Sigma^{ y  x} } - \bs{\Sigma^{ \hat{y}  x}}    )  \bs{Q} diag(\bs{ Q k}) \bs{Q}^{-1}
    %diag(\bs{ \tilde k})^{*T} \bs{Q}^*
\end{align}
where $diag(\bs{Q}^{-1} \bs{k})$ is the $n^2 \times n^2$ diagonal matrix with the elements of $\bs{Q}^{-1} \bs{k}$ placed on the diagonal.  The derivatives of the off-diagonal elements are zero (  $\big ( \frac{d \, diag(\bs{Q^{-1} k})}{dt}  \big )_{i,j} = 0 \, \textrm{if} \, i \neq j$ ).

\subsection{Interpreting the gradient descent equations in terms of the dataset structure}

We can now finally link the process of gradient descent with the structure present in the dataset. We can describe both the input-input as the input-output structure of the dataset with singular value decompositions. We can use these singular value decompositions to perform a change of variables. Starting from:
\begin{align}
\label{eq:SVD_sigma_y_x_repeated}
     & \bs{\Sigma^{ y x}} = \bs{U S V^{T}} \\
     & \bs{\Sigma^{ \hat{y} x}} = \bs{U A V^{T}} \\
     & \bs{\Sigma^{ x x}}= \bs{V \overline{\Sigma^{ x x}} V^{T}}
\end{align}
we transform $\bs{W}$ to $\ol{\bs{W}}$ and $\bs{dbc(K)}$ to $\ol{\bs{dbc(K)}}$: 
\begin{align}
   & \bs W = \bs{ U \overline{W} R^{-1}} \\
   & \iff \bs{\ol{W}} = \bs{ U^{T}} \bs W \bs{R} 
\end{align}
where $\bs{R}$ is an arbitrary invertible matrix, and 
\begin{align}
   & \bs{dbc(K)} = n \bs{Q}^{-1} diag(\bs{Qk})  \bs{Q} =  \bs{R}  \ol{\bs{dbc(K)}}  \bs{V}^{T} \\
   & \iff \ol{\bs{dbc(K)}} = \bs{R}^{-1} \bs{dbc(K)} \bs{V}  = n \bs{R}^{-1}  \bs{Q}^{-1} diag(\bs{Qk})  \bs{Q}  \bs{V}\\
    & \iff \ol{\bs{dbc(K)}} = n \bs{R}^{-1}  \bs{Q} diag(\bs{Q^{-1}k})  \bs{Q}^{-1}  \bs{V} 
\end{align}
where in the last step we used that $\bs{dbc(K)}$ is a real valued matrix, thus $\bs{dbc(K)} = \bs{dbc(K)}^*$, and that $\bs{Q}$ and $\bs{Q}^{-1} (= \bs{Q}^{*T})$ are unitary and symmetric.\\

We perform this change of variables such that we can describe the network, given by the product $\bs{W} \bs{dbc(K)}$, in terms of the singular vectors of the dataset given by $\bs{U}$ and $\bs{V}$. With this change of variables, we indeed have (compare to eq.\ref{eq:SVD_sigma_y_x_repeated}) :
\begin{equation}
    \bs{W}  \bs{dbc(K)} = \bs{ U \overline{W} R^{-1}} \bs{R}  \ol{\bs{dbc(K)}}  \bs{V}^{T} = \bs{ U} \bs{\ol{W}}  \, \bs{\ol{dbc(K)}}  \bs{V}^{T}.
\end{equation}
The matrices $\bs{\ol{W}}$ and  $\bs{\ol{dbc(K)}} $ are thus defined up to an invertible matrix; but their product remains the same. If the $p \times n^2$ product $\bs{\ol{W}} \bs{\ol{dbc(K)}}$ would be rectangular diagonal with $p$ real, positive entries, we would have a singular value decomposition. Moreover, this singular value decomposition would be the same as that of the input-output relationship in the dataset (eq.\ref{eq:SVD_sigma_y_x_repeated}), only with different singular values.  \\

We can also use this change of variables to describe the complete input-predicted output relation $\bs{\Sigma^{\hat y  x}}$ (eq.  \ref{eq:sigma_y_hat_x}):
\begin{align}
\bs{\Sigma^{\hat y x}}  & = \bs{W} \bs{dbc(K)}   \bs{\Sigma^{x  x}} \\
& = \bs{ U \ol{W}  \ol{\bs{dbc(K)}}  \ol{\Sigma^{ x x}}  V}^{T} \\
& = \bs{ U A  V}^{T}  
\end{align}
with $\bs{A}$:
\begin{equation}
    \bs{A} = \bs{\ol{W}}  \ol{\bs{dbc(K)}}  \bs{\ol{\Sigma^{ x x}}} = n \bs{\ol{W}} \bs{R}^{-1}  \bs{Q} diag(\bs{Q^{-1}k})  \bs{Q}^{-1}  \bs{V}  \bs{\ol{\Sigma^{ x x}}  }
\end{equation}
where again, if $\bs{A}$ would be rectangular diagonal with $p$ real, positive entries, this would be an SVD with the same singular vectors as the input-(ground truth) output relationship of the dataset $\bs{\Sigma^{y x}}$. In that case, we will call the matrix $\bs{A}$ the matrix of effective singular values. Note that in practice, we can compute $\bs{A}$ irrespective of the internal details of the network: we only need predictions $\bs{\hat y}$ for samples $\bs{x}$ to compute $\bs{A} = \bs{U^{T}} \bs{\Sigma^{\hat y x}} \bs{V}$. The evolution of $\bs{A}$ during training thus essentially coincides with the evolution of the predictions of the network. \\
We are now ready to rewrite our system of differential equations of gradient descent using the new variables:

\begin{align}
       & \frac{1}{n \lambda}  \big ( \frac{d \, diag(\bs{Q^{-1} k})}{dt}  \big )_{j,j} = \bs Q^{-1}_{j,:} \bs{W}^T\big(  \bs{\Sigma^{ y  x}} - \bs{\Sigma^{ \hat{y}  x}}    )  \bs Q^T_{:,j}  \\
        & \iff \frac{1}{n \lambda}  \big ( \frac{d \, diag(\bs{Q^{-1} k})}{dt}  \big )_{j,j} = \bs Q^{-1}_{j,:} \bs{R}^{-*T} \bs{\ol{W}} ^{*T} \bs{U}^{T} \bs{U}  \big( \bs{S} - \bs{A}     )  \bs{V}^{T}  \bs Q^T_{:,j}
\end{align}

\begin{align}
     & \frac{1}{n\lambda} \bs{U}^{T} \big( \frac{d \bs W }{dt}  \big) \bs{R} = \bs{U}^{T} \big(  \bs{\Sigma^{ y  x} } - \bs{\Sigma^{ \hat{y}  x}}    )  \bs{Q}^T diag(\bs{ Q k}) \bs{Q}^{-1} \bs{R}  \\
        & \iff  \frac{1}{n\lambda} \bs{U}^{T} \big( \frac{d \bs W }{dt}  \big) \bs{R} = \bs{U}^{T} \bs{U}  \big( \bs{S} - \bs{A}     )  \bs{V}^{T}   \bs{Q}^T diag(\bs{ Q k}) \bs{Q}^{-1} \bs{R} 
\end{align}

We thus arrive at the coupled system of equations:
\begin{align}
& \frac{1}{n \lambda}  \big ( \frac{d \, diag(\bs{Q^{-1} k})}{dt}  \big )_{j,j} = \bs Q^{-1}_{j,:} \bs{R}^{-*T} \bs{\ol{W}} ^{*T}   \big( \bs{S} - \bs{A}     )  (\bs{Q} \bs{V})^T_{:,j} \\
& \frac{1}{n\lambda}  \big( \frac{d \bs{\ol{W}} }{dt}  \big) =   \big( \bs{S} - \bs{A}     )  (\bs{Q} \bs{V})^T  diag(\bs{ Q k}) \bs{Q}^{-1} \bs{R}
\end{align}

Which are the equations mentioned in the main paper, Eqs. \ref{eq:update_diag_Qk_transformed} and \ref{eq:update_W_bar}.

\section{Sums of Cosines Datasets}\label{sup_sec:sums_of_cosines}

In the theoretical derivations, we will often use a 'sums of cosines dataset'. This dataset consists of a sum of pure 2D cosines, where the frequencies of those cosines are not shared between classes: 
\begin{equation}
    X^{(c)}_{l,m} =  \sum_{j \in \sigma^{(c)}}  b_{j} \cos( 2 \pi \frac{\mu l}{n} +   2 \pi \frac{\nu m}{n}+ \delta_j),
    \label{eq:sums_of_cosines}
\end{equation}
 where $\sigma^{(c)}$  is a set, belonging to class $c$, of vec-2D indices $j$ that map to pairs of horizontal ($\mu$) and vertical  ($\nu$) frequency indices. $b_j$ is the corresponding amplitude, and $\delta_j$ is the corresponding phase. We here thus consider the case where the sets $\sigma^{(c)}$ are disjoint. In the special case where frequencies are not shared (between modes, see next) we can derive analytical results. \\

 In this case the right singular vectors $\bs{\phi}^{(\alpha)}$ are given by:
 \begin{align}
\bs{\phi}^{(\alpha)}_i &= \frac{1}{\sqrt{\sum_{j \in \sigma^{(\alpha)}} b_j^2}} \frac{n}{\sqrt{2}} \sum_{j \in \sigma^{(\alpha)}}  b_j \cos( 2 \pi \frac{\mu l}{n} +   2 \pi \frac{\nu m}{n}+ (\delta_{\phi})_j), 
\label{eq:sums_of_cosines_phi}
\end{align}
i.e., again the class vectors normalised and sorted according to singular value: the mode indices $\alpha$ form a permutation of the class indices $c$ based on the singular values. Here $\mu = div(j,n), \, \nu = mod(j,n)$ are the frequency indices and  $l = div(i,n), \, m = mod(i,n)$ are the pixel indices. There is a subtlety regarding the phase $(\delta_{\phi})_j$: the SVD is defined up to a minus sign, i.e., if both the left and right singular vector obtain a minus sign, the total mode remains unaltered. Therefore the phase $(\delta_{\phi})_j$ could be the same as the original phase  $(\delta_{\phi})_j= \delta_j $, or a possible minus sign could be absorbed in the phase as $(\delta_{\phi})_j= \delta_j + \pi$. We can furthermore note that the sets \textbf{$\sigma^{\alpha}$}, where each set corresponds to a set \textbf{$\sigma^{(c)}$},  are also disjoint. Formally, this means that given an index $j$, $|(\bs{Q} \bs{\phi}^{\alpha})_j|$ is non-zero for only one mode $\alpha$. The latter property will be crucial in the derivations below. We will also still assume that $\bs{\Sigma^{xx}} = \bs{V} \bs{\ol{\Sigma^{xx}}}\bs{V^T}$ with  $\bs{\ol{\Sigma^{xx}}}$ diagonal. \\

\section{Minimal Norm Weights}

It is well-known that, when gradient descent is used to train a linear neural network from near-zero initial conditions in the over-parametrized regime, the final network parameters will have minimal norm (for a derivation with MSE loss, see e.g., \cite{bach2023principles}, chapter 11). I.e., among all the possible combinations of network weights that implement the desired input-output map, gradient descent leads to the network weights that have the smallest norm.  This is called the implicit regularization of gradient descent. The minimal norm solution for the two-layer linear CNN with a single kernel is given by the following constrained minimization problem:
\begin{equation}\label{eq:min_norm}
    \text{min}_{\bs{W}, \bs{dbc(K)}} \Big( ||\bs{W}||^2_{F} +  ||\bs{dbc(K)}||^2_{F}  \Big)
\end{equation}
subject to:
\begin{equation}\label{eq:constrained_min_norm}
    \bs{W} \bs{dbc(K)}  \bs{\Sigma^{xx}}=  \bs{U} \bs{S} \bs{V^T}
\end{equation}
where $||\bs{B}||_{F}$ denotes the Frobenius norm of the $l \times m$ matrix $\bs{B}$, in essence $||\bs{B}||_{F} = \sqrt{\sum^l_{i} \sum^m_j |B_{ij}|}$.

\subsection{Exact Solutions for Sum of Cosines Datasets}\label{sec:lagrange_sol}

 We here derive what those final solutions look for a sums of cosines dataset. We first define three new $n^2 \times 1$ vectors, $\bs{\delta}_\phi$, and  $\bs{\delta}_k$ and $\bs{\delta}_w$.  These vectors contain the phases (or angles)  associated to the complex-valued vec-2D DFT of the singular values, of the kernel and of the weights, respectively. The definition of $\bs{\delta}_k$ is straightforward:
\begin{equation}
    \bs{\delta}_{\bs{k}} = angle(\bs{Qk}).
\end{equation}
The definition of $\bs{\delta}_\phi$ and $\bs{\delta}_w$ is more subtle. The $p \times n^2$ matrix $\bs{W}$ has $p$ rows, and there are also $p$  singular vectors $\bs{\phi}^\alpha$ ($\alpha \in \{0, \cdots, p-1\}$) of dimension $n^2 \times 1$. For every index $j$, there are thus $p$ complex numbers, e.g.,  $(\bs{Q} \bs{\phi}^0)_j$, $(\bs{Q} \bs{\phi}^1)_j$, etc. However, when we use a dataset with disjoint sets of frequencies, we have by design a non-zero value  $(\bs{Q} \bs{\phi}^\alpha)_j$ for only one single mode index $\alpha$ per frequency index $j$. We select the phase of this value to be the $j^{th}$ element of the vector  $\bs{\delta}_\phi$. We thus have:
\begin{align}
\begin{cases}
         (\bs{\delta}_{\phi})_j =  angle(\bs{Q} \bs{\phi}^\alpha)_j, &  j \rightarrow \alpha, \\
         (\bs{\delta}_{\phi})_j =  0, & \text{freq. $j$ not present in singular vectors.} \\
\end{cases}
\end{align}
where again $j \rightarrow \alpha$ denotes that the frequency $j$ is used in the sum of cosines making up the singular vector $\bs{\phi}^\alpha$. We define $\bs{\delta}_w$ in a similar way:
\begin{align}
\begin{cases}
        (\bs{\delta}_{w})_j =  angle( (\bs{Q} \bs{W}^T)_{j,\alpha}), &  j \rightarrow\alpha, \\
        (\bs{\delta}_{w})_j =  0, &   \text{freq. $j$ not present in singular vectors.}
\end{cases}
\end{align}
Note that because the kernel, the weights and the singular vectors are all real-valued, we have $(\bs{\delta}_{\phi})_{j_{symm}} = - (\bs{\delta}_{\phi})_{j} $, $(\bs{\delta}_{k})_{j_{symm}} = - (\bs{\delta}_{k})_{j} $, and  $(\bs{\delta}_{w})_{j_{symm}} = - (\bs{\delta}_{w})_{j} $.\\

We are now ready to state what the final, minimal norm solutions look like. A kernel $\bs{K}$ and the weight matrix $\bs{W}$ correspond to a minimal norm solution, if and only if: \\
\begin{equation}\label{eq:S_min_norm}
S_{\alpha,\alpha} = n \sum\limits_{j \in \sigma^{\alpha}_{symm}}  |(\bs{Q} \bs{\phi}^\alpha)_j| \, \, |(\bs{Qk})_j|^2 \, \, \ol{\Sigma}_{\alpha,\alpha}, 
\end{equation}
with $\sigma_{symm}^{\alpha }$ is the original set of indices $j$ in $\sigma^{\alpha }$ together with all their corresponding symmetric indices $j_{symm}$, and
\begin{equation}
    \bs{W}=\bs{U}  \bs{\Omega} \bs{\Theta}_{w} \bs{Q},
\end{equation}
where $\bs{\Omega}$ is a $p \times n^2$ matrix defined by: 
\begin{equation}
    \begin{cases}
\bs{\Omega}_{\alpha, j} = |\bs{(Q k)}_j|,   \, \,  \bs{\Omega}_{\alpha, j_{symm}} = |\bs{(Q k)}_j| &  j \rightarrow \alpha,  \\
\bs{\Omega}_{\alpha, j} = 0  & else,  \\
\end{cases}
\label{eq:W_bar_sol_real}
\end{equation} 
and 
\begin{equation}
     \bs{\Theta}_{w}  = diag(\bs{e}^{i\bs{\delta}_w}). 
\end{equation}
With, for all $j \in \sigma_{symm}^{\alpha}$: %$j \in \{0, \cdots, n^2 -1 \}$:
\begin{align}
& \,(\bs{\delta}_k)_j + (\bs{\delta}_w)_j  = - (\bs{\delta}_\phi)_j .
\end{align}

From these minimal norm solutions, we can then also conclude (cfr. Eq.  \ref{eq:transform_W}):
\begin{align}
    & \bs{R} = \bs{Q}^{-1},\\
   & \bs{\ol{W}}= \bs{\Omega}\bs{\Theta_w}.
\end{align}

Note that the only constraint on the Fourier spectrum of the kernel is given by Eq. \ref{eq:S_min_norm}; the kernel is not uniquely defined. We show in the main paper that the final coefficients $|(Qk)_j|$ are influenced or determined through a winner-takes-all process that takes place during training with gradient descent. This effect can only be derived from the gradient equations themselves, however.

\textbf{Proof overview:} \\
We start from $\bs{dbc(K)}= n \bs{Q}^{-1} diag(\bs{Qk}) \bs{Q}$, with $\bs{k} = vec(\bs{K})$ an arbitrary, real-valued kernel, and from $\bs{W} = \bs{U} \bs{P}$ where $\bs{P}$ is an arbitrary $n \times n$ real-valued matrix, such that $\bs{W}$ is an arbitrary real-valued weight matrix. We then use the method of Lagrange multipliers on the constrained optimization problem given by Eq. \ref{eq:min_norm} and Eq. \ref{eq:constrained_min_norm} to find a system of equations involving $\bs{k}$, $\bs{P}$, and the given matrices $\bs{U}$,  $\bs{V}$, $\bs{S}$, $\bs{Q}$ and $\bs{\Sigma}^{xx}$. This system of equations captures the relationships between these matrices when $\bs{dbc(K)}$ and $\bs{W}$ form a minimal norm solution. When frequencies are not shared between modes, i.e., when for every frequency index $j$ only one value $\bs{Q}_{j,:} \bs{V}_{:,\alpha}$ is significant, the solutions to these equations are as described above. 

\subsection{Detailed Proof}

Let $\bs{dbc(K)}= n \bs{Q}^{-1} diag(\bs{Qk}) \bs{Q}$, with $\bs{k} = vec(\bs{K})$ an arbitrary, real-valued kernel, and let $\bs{W} = \bs{U} \bs{P}$ where $\bs{P}$ is an arbitrary $n \times n$ real-valued matrix, such that $\bs{W}$ is an arbitrary real-valued weight matrix. We first reformulate the constrained minimization problem Eq. \ref{eq:min_norm} using properties of the Frobenius norm. The Frobenius norm can be expressed in function of singular values:\scalebox{0.8}{ $||\bs{B}||_{F} =   \sqrt{\sum^{min(l,m)}_{i} \sigma_i(\bs{B})^2 } $} where $\sigma_i(\bs{B})$ are the singular values of $\bs{B}$. We can make use of the latter property of the Frobenius norm to explicitly include the constraint on the network structure, given by the eigendecomposition $\bs{dbc(K)} = n \bs{Q}^{-1} diag(\bs{Qk}) \bs{Q}$. The singular values of $\bs{dbc(K)}$ are the square roots of the eigenvalues of $\bs{dbc(K)}^T \bs{dbc(K)} = n^2 \bs{Q}^{-1} diag(\bs{Q^*k}) diag(\bs{Qk}) \bs{Q}$. The singular values of $\bs{dbc(K)}$ are thus given by  $n |(\bs{Qk})_j|$. The Frobenius norm can also be expressed as a trace, \scalebox{0.8}{$||\bs{B}||_{F} = \sqrt{tr(\bs{B}^{*T} \bs{B})}  $}. Therefore $||\bs{W}||^2_{F}$ can be expressed as $tr \big( \bs{P}^{T} \bs{U}^T \bs{U} \bs{P} \big) = tr \big( \bs{P}^{T} \bs{P} \big)$, given that $\bs{P}$ is real and $\bs{U}$ is real and orthogonal.  \\
We thus reformulate the constrained minimization problem (Eq. \ref{eq:min_norm} and \ref{eq:constrained_min_norm}) as:
\begin{equation}
    \text{min}_{\bs{P} , \bs{Qk}} \Big( tr \big( \bs{P}^{*T}  \bs{P} \big)  +   \sum_i^{n^2} {|(\bs{Qk})_i|^2} \Big),
    \label{eq:constr_op_target}
\end{equation}
subject to 
\begin{equation}
    n \bs{P} \bs{Q}^{-1} diag(\bs{Qk}) \bs{Q} \bs{V} \bs{\ol{\Sigma^{xx}}}  = \bs{S}. 
     \label{eq:constr_op_constraint}
\end{equation}

We can make use of Lagrange multipliers to solve the constrained optimization problem given by Eqs. \ref{eq:constr_op_target} and \ref{eq:constr_op_constraint}. We denote the multipliers by the $p \times n^2$ matrix $\bs{\Lambda}$, and the Lagrangian $\mathcal{L}$ is then given by:
\begin{equation}
    \mathcal{L} = tr \Big(\bs{P}^{T} \bs{P}  \Big) +  \sum_j^{n^2} |(\bs{Qk})_j|^2 + tr \Big( \bs{\Lambda}^T \big( n \bs{P} \bs{Q}^{-1} diag(\bs{Qk}) \bs{Q} \bs{V} \bs{\ol{\Sigma^{xx}}} - \bs{S}   \big) \Big)
\end{equation}
To find the minimal norm solution, we compute the derivatives with respect to the matrix $\bs{P}$, the values $(Qk)_j$ and the matrix $\bs{\Lambda}$. These derivatives are zero at the minimum. \\

$\frac{d \mathcal{L}}{d \bs{P}}$:\\

To compute the derivative with respect to the matrix $\bs{P}$, we make use of the cyclic property of the trace function, and the fact that $\frac{d \, tr(\bs{B}\bs{C})}{d \bs{B}} = \bs{C}^T$. We obtain:
\begin{align}
\frac{d \mathcal{L}}{d \bs{P}} &= \frac{d }{d \bs{P}} \Big(  tr\big( \bs{P} \bs{P}^{T}  \big)  + tr \big(  \bs{P} \bs{Q}^{-1} diag(\bs{Qk}) \bs{Q} \bs{V} \bs{\ol{\Sigma^{xx}}} \bs{\Lambda}^T  \big)  \Big) \\
&= \bs{P} + \bs{\Lambda} \bs{\ol{\Sigma^{xx}}}^T \bs{V}^T  \bs{Q}^T diag(\bs{Qk}) \bs{Q^{-T}} 
\end{align}

$\frac{d \mathcal{L}}{d \bs{(Qk)}_j}$:\\

To calculate this derivative, we will make use of the following properties:
\begin{equation}
\big(diag(\bs{Qk}) \bs{Q} \big)_{l,m} =   \sum_i diag(\bs{Qk})_{l,i} \bs{Q}_{i,m} = diag(\bs{Qk})_{l,l} \bs{Q}_{l,m}
\end{equation}
yielding
\begin{equation}
    (\bs{Q}^{-1} diag(\bs{Qk}) \bs{Q})_{i,m} =  \sum_l  (\bs{Q}^{-1}_{i,l} \big( diag(\bs{Qk}) \bs{Q} \big)_{l,m}  =  \sum_l  \bs{Q}^{-1}_{i,l} diag(\bs{Qk})_{l,l} \bs{Q}_{l,m}
\end{equation}
and, for any matrix $\bs{B}$ and  $\bs{C}$, 
\begin{equation}
    tr(\bs{B}\bs{C}) = \sum_i\bs{B}_{i,:} \bs{C}_{:,i}
\end{equation}
since the trace operation is the sum of diagonal elements. Moreover, we can compute the derivative of a function $f$ with respect to a complex number $z = z_\mathbb{R} + i z_\mathbb{I}$ as $\frac{d f}{dz } = \frac{1}{2} (\frac{d f}{dz_\mathbb{R}} - i  \frac{d f}{dz_\mathbb{I}})$ (see, e.g., \cite{gunning1965analytic}).  
The derivative of the third term is then given by:
\begin{align}
    & \frac{d}{d \, diag(\bs{Qk})_{j,j}} \Big(  tr \big(  \bs{P} \bs{Q}^{-1} diag(\bs{Qk}) \bs{Q} \bs{V} \bs{\ol{\Sigma^{xx}}} \bs{\Lambda}^T  \big)   \Big) \\
    & = \frac{d}{d \, diag(\bs{Qk})_{j,j}} \Big( tr \big( \bs{P}  \sum_l \bs{Q}_{:,l}^{-1} diag(\bs{Qk})_{l,l} \bs{Q}_{l,:}  \bs{V} \bs{\ol{\Sigma^{xx}}} \bs{\Lambda}^T    \big)    \Big) \\
    & = \frac{d}{d \, diag(\bs{Qk})_{j,j}} \Big( \sum_m \bs{P}_{m, :}   \sum_l \bs{Q}_{:,l}^{-1} diag(\bs{Qk})_{l,l} \bs{Q}_{l,:}  \bs{V} \bs{\ol{\Sigma^{xx}}}  \bs{\Lambda}^T_{:,m}      \Big) \\
    &= \frac{1}{2} ( \sum_m \bs{P}_{m, :}  \bs{Q}_{:,j}^{-1}  \bs{Q}_{j,:}  \bs{V} \bs{\Lambda}^T_{:,m}  - i^2 \sum_m \bs{P}_{m, :}  \bs{Q}_{:,j}^{-1}  \bs{Q}_{j,:}  \bs{V} \bs{\Lambda}^T_{:,m})  \\
    &= \sum_m \bs{P}_{m, :}  \bs{Q}_{:,j}^{-1}  \bs{Q}_{j,:}  \bs{V} \bs{\ol{\Sigma^{xx}}}  \bs{\Lambda}^T_{:,m} \\
    &= tr( \bs{P}  \bs{Q}_{:,j}^{-1}  \bs{Q}_{j,:}  \bs{V} \bs{\ol{\Sigma^{xx}}}  \bs{\Lambda}^T   )
\end{align}
This yields, for the total derivative $\frac{d \mathcal{L}}{d \bs{(Qk)}_j}$:
\begin{align}
    \frac{d \mathcal{L}}{d \bs{(Qk)}_j} &= \frac{1}{2} \big(2 \bs{(Qk)}_{\mathbb{R} j} -2i\bs{(Qk)}_{\mathbb{I}j} \big) + tr( \bs{P}  \bs{Q}_{:,j}^{-1}  \bs{Q}_{j,:}  \bs{V} \bs{\ol{\Sigma^{xx}}}  \bs{\Lambda}^T   ) \\
    &= (\bs{Qk})^*_j + tr( \bs{P}  \bs{Q}_{:,j}^{-1}  \bs{Q}_{j,:}  \bs{V} \bs{\ol{\Sigma^{xx}}}  \bs{\Lambda}^T   )
\end{align}

$\frac{d \mathcal{L}}{d \bs{\Lambda}} $:\\

Finally, we compute $\frac{d \mathcal{L}}{d \bs{\Lambda}}$:
\begin{align}
    \frac{d \mathcal{L}}{d \bs{\Lambda}}  &= \frac{1}{2} \big(  \frac{d \mathcal{L}}{d \bs{\Lambda_\mathbb{R}}} - i  \frac{d \mathcal{L}}{d \bs{\Lambda_\mathbb{I}}} \big) \\
    &= \frac{1}{2} \Big( \big( n \bs{P} \bs{Q}^{-1} diag(\bs{Qk}) \bs{Q} \bs{V} \bs{\ol{\Sigma^{xx}}} - \bs{S}   \big) -i^2 \big( n \bs{P} \bs{Q}^{-1} diag(\bs{Qk}) \bs{Q} \bs{V} \bs{\ol{\Sigma^{xx}}} - \bs{S}   \big)   \Big)  \\
   & = n \bs{P} \bs{Q}^{-1} diag(\bs{Qk}) \bs{Q} \bs{V} \bs{\ol{\Sigma^{xx}}} - \bs{S}   
\end{align}

The complete system of equations is then given by:

\begin{align}
\begin{cases}
\frac{d \mathcal{L}}{d \bs{P}} = 0 \\
\frac{d \mathcal{L}}{d \bs{(Qk)}_j} = 0 \\
\frac{d \mathcal{L}}{d \bs{\Lambda}} = 0\\
\end{cases}
\iff
& \begin{cases}
\bs{P} + \bs{\Lambda} \bs{\ol{\Sigma^{xx}}}^T \bs{V}^T  \bs{Q}^T diag(\bs{Qk}) \bs{Q^{-T}}  = 0 \\
(\bs{Qk})^*_j + tr( \bs{P}  \bs{Q}_{:,j}^{-1}  \bs{Q}_{j,:}  \bs{V} \bs{\ol{\Sigma^{xx}}} \bs{\Lambda}^T   ) = 0 \\
 n \bs{P} \bs{Q}^{-1} diag(\bs{Qk}) \bs{Q} \bs{V} \bs{\ol{\Sigma^{xx}}} - \bs{S} = 0 
 \label{eq:lagrange_system}
\end{cases}
\end{align}

From the first equation of eqs. \ref{eq:lagrange_system} we obtain:
\begin{align}
&\bs{P} = - \bs{\Lambda} \bs{\ol{\Sigma^{xx}}}^T \bs{V}^T  \bs{Q}^T diag(\bs{Qk}) \bs{Q^{-T}}  \\
\iff &\bs{P} = - \bs{\Lambda} \bs{\ol{\Sigma^{xx}}}^T \bs{V}^T  \bs{Q}^{-T} diag(\bs{Qk})^* \bs{Q}^T
\end{align}
where we used that $\bs{dbc(K)}$ is real-valued. We can plug this in the second equation of the system of eqs. \ref{eq:lagrange_system} :
\begin{align}
    & (\bs{Qk})^*_j + tr( \bs{P}  \bs{Q}_{:,j}^{-1}  \bs{Q}_{j,:}  \bs{V} \bs{\ol{\Sigma^{xx}}} \bs{\Lambda}^T   ) = 0 \\
   \iff  & (\bs{Qk})^*_j +  tr( - \bs{\Lambda} \bs{\ol{\Sigma^{xx}}}^T \bs{V}^T  \bs{Q}^{-T} diag(\bs{Qk})^* \bs{Q}^T \bs{Q}_{:,j}^{-1}  \bs{Q}_{j,:}  \bs{V} \bs{\ol{\Sigma^{xx}}} \bs{\Lambda}^T   ) = 0 \\
     \iff  &  (\bs{Qk})^*_j +  tr( - \bs{\Lambda} \bs{\ol{\Sigma^{xx}}}^T \bs{V}^T  \bs{Q}^{-T}_{:,j} diag(\bs{Qk})^*_{j,j} \bs{Q}_{j,:}  \bs{V} \bs{\ol{\Sigma^{xx}}} \bs{\Lambda}^T   ) = 0 
\end{align}
If the frequency indexed by $j$ is not present in any mode, i.e., $(\bs{Q} \bs{V})_{j,\alpha}  = 0$ for all mode indices $\alpha$, then we obtain:
\begin{equation}
     (\bs{Qk})^*_j = (\bs{Qk})_j = 0
\end{equation}
If the frequency indexed by $j$ is present in mode $\alpha$, then  $(\bs{Q} \bs{V})_{j,\alpha}  \neq 0$ for a single mode $\alpha$ (by assumption). We then obtain:
\begin{align}
 (\bs{Qk})^*_j +  tr( - \bs{\Lambda} \bs{\ol{\Sigma^{xx}}}^T_{:, \alpha} (\bs{V}^T  \bs{Q}^{-T})_{\alpha,j} diag(\bs{Qk})^*_{j,j} (\bs{Q} \bs{V})_{j,\alpha}  \bs{\ol{\Sigma^{xx}}}_{\alpha, :}\bs{\Lambda}^T   ) = 0
\end{align}
and because $\bs{\ol{\Sigma^{xx}}}$ is diagonal (by assumption):
\begin{align}
 &(\bs{Qk})^*_j +  tr( - \bs{\Lambda}_{:,\alpha}\bs{\ol{\Sigma^{xx}}}^T_{\alpha, \alpha} (\bs{V}^T  \bs{Q}^{-T})_{\alpha,j} diag(\bs{Qk})^*_{j,j} (\bs{Q} \bs{V})_{j,\alpha}  \bs{\ol{\Sigma^{xx}}}_{\alpha, \alpha}\bs{\Lambda}^T_{\alpha, :}   ) = 0 \\
 \iff & (\bs{Qk})^*_j  + tr( - \bs{\Lambda}_{:,\alpha}\bs{\ol{\Sigma^{xx}}}^T_{\alpha, \alpha} ( \bs{Q}\bs{V} )^{*T}_{\alpha,j}  diag(\bs{Qk})^*_{j,j} (\bs{Q} \bs{V})_{j,\alpha}  \bs{\ol{\Sigma^{xx}}}_{\alpha, \alpha}\bs{\Lambda}^T_{\alpha, :}   ) = 0  \\
  \iff & (\bs{Qk})^*_j  + tr( - \bs{\Lambda}_{:,\alpha}\bs{\ol{\Sigma^{xx}}}^T_{\alpha, \alpha}  diag(\bs{Qk})^*_{j,j} |(\bs{Q} \bs{V})_{j,\alpha}|^2  \bs{\ol{\Sigma^{xx}}}_{\alpha, \alpha}\bs{\Lambda}^T_{\alpha, :}   ) = 0 \\
    \iff & (\bs{Qk})^*_j  -  (\bs{Qk})^*_j  \bs{\ol{\Sigma^{xx}}}_{\alpha, \alpha}^2 |(\bs{Q} \bs{V})_{j,\alpha}|^2   tr( \bs{\Lambda}_{:,\alpha}\bs{\Lambda}^T_{\alpha, :}   ) = 0 \\ 
    \iff & \bs{\Lambda}^T_{\alpha, :}  \bs{\Lambda}_{:,\alpha} = \bs{\ol{\Sigma^{xx}}}_{\alpha, \alpha}^{-2} |(\bs{Q} \bs{V})_{j,\alpha}|^{-2}  
\end{align}

This yields:
\begin{equation}
    \bs{\Lambda}_{:, \alpha} =  \bs{\tilde{R}}_{:,\alpha}\bs{\ol{\Sigma^{xx}}}_{\alpha, \alpha}^{-1} |(\bs{Q} \bs{V})_{j,\alpha}|^{-1} 
\end{equation}
with $\bs{\tilde{R}}$ a $p\times p$ orthogonal matrix. \\

We now propose to decompose $\bs{P}$ as $\bs{P} = \bs{\Omega} \bs{\Theta}_{w} \bs{Q}$. We can use the obtained equations to determine the shape of $\bs{\Omega} $ and $\bs{\Theta}_{w}$ corresponding to a minimal norm solution. From the first equation:

\begin{align}
& \bs{P} = - \bs{\Lambda} \bs{\ol{\Sigma^{xx}}}^T \bs{V}^T  \bs{Q}^{-T} diag(\bs{Qk})^* \bs{Q}^T \\
\iff & \bs{\Omega} \bs{\Theta}_{w} = - \bs{\Lambda} \bs{\ol{\Sigma^{xx}}}^T \bs{V}^T  \bs{Q}^{-T} diag(\bs{Qk})^* \\
\iff & (\bs{\Omega} \bs{\Theta}_{w})_{:,j} = - \bs{\Lambda}_{:,\alpha} \bs{\ol{\Sigma^{xx}}}^T_{\alpha,\alpha}  (\bs{Q} \bs{V})^{*T}_{\alpha,j} diag(\bs{Qk})^*_{j,j} \\
\iff & (\bs{\Omega} \bs{\Theta}_{w})_{:,j} = - \bs{\tilde{R}}_{:,\alpha} \bs{\ol{\Sigma^{xx}}}_{\alpha, \alpha}^{-1} |(\bs{Q} \bs{V})_{j,\alpha}|^{-1}  \bs{\ol{\Sigma^{xx}}}^T_{\alpha,\alpha}  (\bs{Q} \bs{V})^{*T}_{\alpha,j} diag(\bs{Qk})^*_{j,j} \\
\iff & (\bs{\Omega} \bs{\Theta}_{w})_{:,j} = - \bs{\tilde{R}}_{:,\alpha}  |(\bs{Q} \bs{V})_{j,\alpha}|^{-1} e^{- i (\bs{\delta_{\phi}})_j} |(\bs{Q} \bs{V})_{\alpha,j}|  e^{- i (\bs{\delta_k})_j} |(\bs{Qk})_{j}| \\
\iff & \bs{\Omega}_{:,j}  (\bs{\Theta}_{w})_{j,j}= - \bs{\tilde{R}}_{:,\alpha}   e^{- i (\bs{\delta_{\phi}})_j} e^{-i (\bs{\delta_k})_j} |(\bs{Qk})_{j}| 
\end{align}
Thus we obtain, for the real-valued matrix $\bs{\Omega}$:
\begin{equation}
     \bs{\Omega}_{:,j} =  - \bs{\tilde{R}}_{:,\alpha} |(\bs{Qk})_{j}| 
\end{equation}
and for the diagonal matrix with associated phases:
\begin{equation}
    (\bs{\Theta}_{w})_{j,j} = e^{ i (\bs{\delta_w})_j }   = e^{ i \big(-(\bs{\delta_{\phi}})_j - (\bs{\delta_k})_j \big)}  ,
\end{equation}
or also
\begin{equation}
    (\bs{\delta_w})_j   = -(\bs{\delta_{\phi}})_j - (\bs{\delta_k})_j .
\end{equation}
Finally, we can use the third equation of the system of eqs. \ref{eq:lagrange_system}, involving the $p \times n^2$ rectangular diagonal matrix $\bs{S}$ with real, positive values on the diagonal of $\bs{S}_{0:p, 0:p}$, and zero elsewhere. Starting with the elements of $\bs{S}_{0:p, 0:p}$:
\begin{align}
\bs{S}_{\beta, \alpha} &= n \bs{\Omega}_{\beta,:} \bs{\Theta}_{w} diag(\bs{Qk}) \bs{Q} \bs{V} \bs{\ol{\Sigma^{xx}}}_{:, \alpha}  \\
& = \sum_j n \bs{\Omega}_{\beta,j} (\bs{\Theta}_{w})_{j,j} diag(\bs{Qk})_{j,j} (\bs{Q} \bs{V})_{j,\alpha}\bs{\ol{\Sigma^{xx}}}_{\alpha, \alpha} \\
& = \sum_j n \bs{\Omega}_{\beta,j} (\bs{\Theta_w})_{j,j} (\bs{\Theta}_{k})_{j,j} (\bs{\Theta}_{\phi})_{j,j} |diag(\bs{Qk})_{j,j}| |(\bs{Q} \bs{V})_{j,\alpha}|\bs{\ol{\Sigma^{xx}}}_{\alpha, \alpha}\\
& = \sum_j n \bs{\Omega}_{\beta,j}  |diag(\bs{Qk})_{j,j}| |(\bs{Q} \bs{V})_{j,\alpha}|\bs{\ol{\Sigma^{xx}}}_{\alpha, \alpha}\\
& = \sum_j  - n \bs{\tilde{R}}_{\beta,\alpha} |(\bs{Qk})_{j}|^2 |(\bs{Q} \bs{V})_{j,\alpha}|\bs{\ol{\Sigma^{xx}}}_{\alpha, \alpha}
\end{align}
If $\beta \neq \alpha$, then $\bs{S}_{\beta, \alpha} =0$. Therefore $\bs{\tilde{R}}_{\beta,\alpha}=0$ if $\beta \neq \alpha$. Given that $\bs{\tilde R}$ is orthogonal, and $\bs{S}_{\alpha, \alpha}$ is positive and real, we have $\bs{\tilde{R}}_{\alpha,\alpha}=-1$. The other elements of $\bs{S}$,  $\bs{S}_{\alpha, i}$ with $i \>= p$, are indeed zero, since $\bs{\ol{\Sigma^{xx}}}_{\alpha, i} =0$.  This completes the proof.\\

For completeness, we also show that $\bs{P}$ is a real-valued matrix:
\begin{align}
    \bs{P}_{l,m} &= (\bs{\Omega} \bs{\Theta}_{w} \bs{Q})_{l,m} \\ &= \sum_j \bs{\Omega}_{l,j} diag(\bs{e}^{i\bs{\delta}_w})_{j,j}  \bs{Q}_{j,m} \\
    &= \sum_{j \in \sigma^{\alpha}} \Big(  \bs{\Omega}_{l,j} diag(\bs{e}^{i\bs{\delta}_w})_{j,j}  \bs{Q}_{j,m}  + \bs{\Omega}_{l,j_{symm}} diag(\bs{e}^{i\bs{\delta}_w})_{j_{symm},j_{symm}}  \bs{Q}_{j_{symm},m}  \Big)\label{eq:P_real}
\end{align}

Given the vec-2D indices and their corresponding pair of horizontal and vertical indices, $j \rightarrow (\mu, \nu)$, $m \rightarrow (a, b)$, we find:

\begin{align}
    n \bs{Q}_{j,m} & = (\bs{F} \otimes \bs{F} )_{ (\mu-1)n + \nu, (a-1)n + b}  \\ &= F_{\mu,a}  F_{\nu,b} \\ &= e^{- \frac{2\pi i }{n}(\mu a + \nu b)}
\end{align}

and for $j_{symm} \rightarrow (n-\mu, n-\nu)$: 
\begin{align}
    n \bs{Q}_{j_{symm},m} &=  F_{n-\mu,a}  F_{n-\nu,b} \\&= e^{- \frac{2\pi i}{n}  ((n-\mu) a + (n-\nu)b)} \\&= e^{-2\pi i  \frac{n}{n} a } e^{-2\pi  i \frac{n}{n} b }  e^{\frac{+2\pi i }{n}  (\mu a + \nu b)} \\
    & = e^{+\frac{2\pi i}{n} (\mu a + \nu b)} \\
    &=  n \bs{Q}^*_{j,m} 
\end{align}

The vector $\bs{\delta}_w$ has the property that $(\bs{\delta}_w)_j = - (\bs{\delta}_w)_{j_{symm}}$. This yields $diag(\bs{e}^{i\bs{\delta}_w})_{j_{symm},j_{symm}} =  diag(\bs{e}^{i\bs{\delta}_w})^*_{j,j}$. Since $\bs{\Omega}$ is real-valued, each term in the summation in Eq.\ref{eq:P_real} is real, and therefore $\bs{P}$ is real.

\section{Silent Alignment, Balanced Weights and WTA Dynamics}\label{sup_sec:align_balanced_WTA}

In the previous section, we derived the shape of the final weights and kernel for datasets consisting of sums of cosines. These were the minimal norm solutions---the kind of network parameters the linear CNN converges to when starting from small, random initial conditions. In the next section, we derive analytic solutions to the evolution of the predictions of the network, for specific datasets and assuming \emph{aligned} and \emph{balanced} initial conditions. With \emph{aligned} initial conditions, we mean initial conditions that render the matrix $\bs{A}$ rectangular diagonal, with positive real, entries from the start. The alignment can be interpreted as follows: if $\bs{A}$ is such a rectangular diagonal matrix, the network parameters are aligned with the statistical dataset structure, in the sense that $\bs{\Sigma^{\hat y  x}}$ and $ \bs{\Sigma^{y  x}}$ have the same singular vectors. With \emph{balanced} we mean that the (Fourier transformed) weight matrix consists of (Fourier transformed) elements of the kernel; implying some balancedness between the two types of layers. We borrowed this term from the theory of fully connected networks \cite{saxe2019mathematical}, but the shapes of the matrix involved for CNN are more complicated, which makes the concept more subtle.

The assumption of aligned and balanced initial conditions turns out to be a good approximation to small, random initial conditions and a sums of cosines dataset: when starting from small, random initial conditions, we see that the network parameters quickly become aligned and balanced. This also happens in a stage-like fashion,  and the alignment associated to a mode occurs before that mode is learned and the loss appreciably decreases. Therefore the dynamics of learning can be approximately described by assuming aligned and balanced conditions from the start. This phenomenon has been rigorously analysed for the case of fully connected networks \cite{atanasov2022neural, braun2022exact}. It has also been shown that the error of this approximation decreases with initialization scale, e.g., it decreases as the standard deviation $\sigma$ used in a gaussian initializer decreases. \\

\begin{figure}[h!]
\centering
\includegraphics{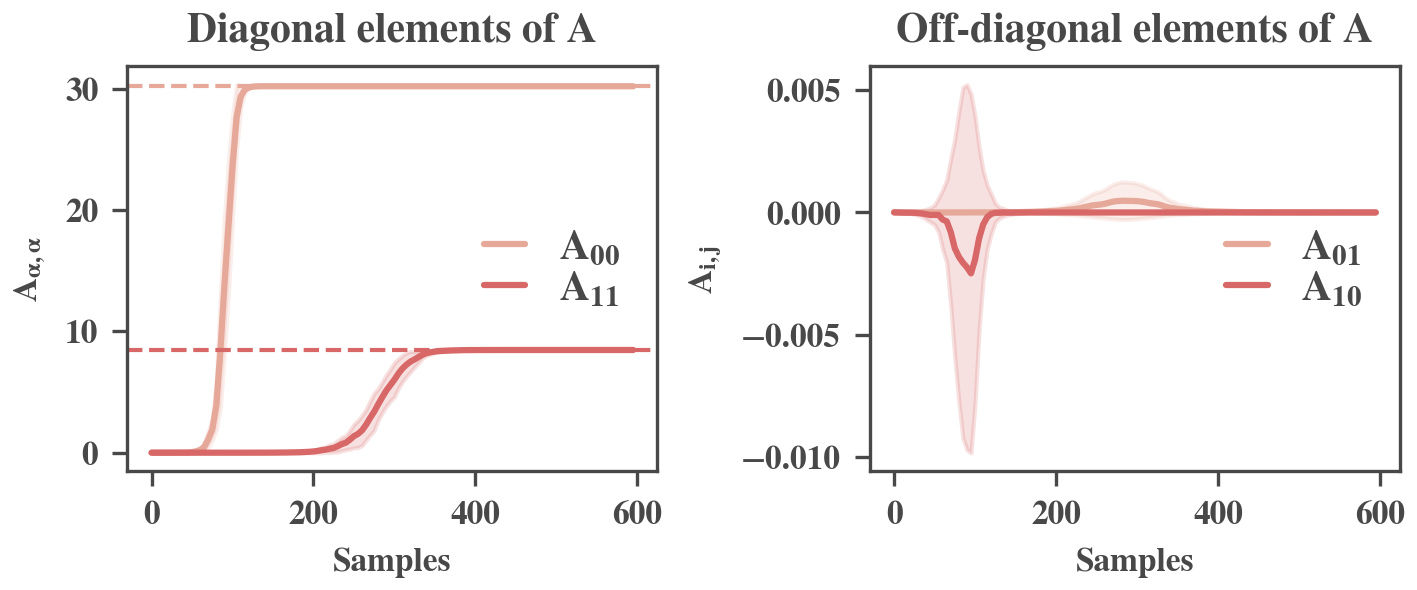}
\caption{ Evolution of the matrix $\bs{A}_{0:2, 0:2}$ during training: diagonal elements (left) and off-diagonal elements (right).  Solid lines: average over trials. Shaded regions: standard deviation over trials. Horizontal, dashed lines: singular values $s_\alpha$. }
\label{fig:evo_A}
\end{figure}

\begin{figure}[h!]
\centering
\includegraphics{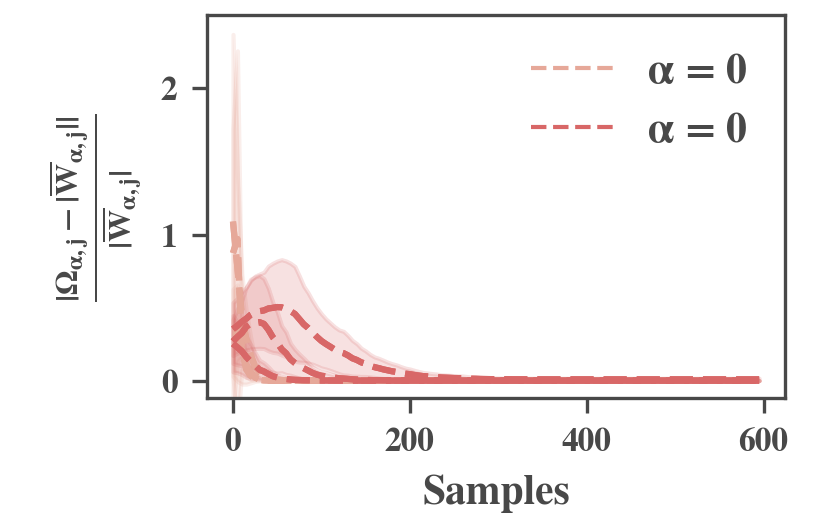}
\caption{ Plot of the evolution of the different fractions $\frac{ | \, |\bs{\ol{W}}_{\alpha, j}| - \bs{\Omega}_{\alpha, j} | }{|\bs{\ol{W}}_{\alpha, j}|}$, a measure of how quickly the network parameters become balanced.  }
\label{fig:dynamics_of_learning_balanced}
\end{figure}

In Fig. \ref{fig:evo_A},  we have plotted, for the sums of cosines dataset, both the evolution of the two diagonal and the two off-diagonal elements of $\bs{A}_{0:2, 0:2}$, averaged over 10 trials using small, random, initial conditions ($\sigma = 0.00001$). As discussed in the main paper, the diagonal values of $\bs{A}$ follow sigmoidal, stage-like trajectories: after some time, they suddenly appear to grow very fast, eventually saturating on their corresponding singular value. I.e.,  $A_{0, 0}$ saturates at the value $s_0$ and $A_{1, 1}$ saturates at the value $s_1$. The off-diagonal elements, on the other hand, remain relatively small (compare the y-axis between the left and right plot) and exhibit transient peaks, before becoming $\approx 0$. \\

Fig.\ref{fig:dynamics_of_learning_balanced} shows how quickly the magnitudes of the elements $\bs{\ol{W}}_{\alpha, j}$ become equal to $\bs{\Omega}_{\alpha, j} = |(\bs{Qk})_j|$, i.e., how quickly the solution becomes balanced. We plotted the evolution of the fraction $\frac{ | \, |\bs{\ol{W}}_{\alpha, j}| - \bs{\Omega}_{\alpha, j} | }{|\bs{\ol{W}}_{\alpha, j}|}$ during training, for all the different frequencies (indexed by $j$) involved in the two modes $\alpha =0$ and $\alpha =1$ of the sums of cosines dataset. $\bs{\ol{W}}$ is computed from $\bs{W}$ using $\bs{R} = \bs{Q}^{-1}$. We see that the balancedness also happens in a stage-like or cascaded fashion: very quickly, the elements corresponding to the first mode become balanced; in a next phase and after the first mode has been learned, the elements corresponding to the second mode also become balanced. \\

For the sums of cosines dataset, we thus quickly arrive at the following shape for the fully connected weight matrix $\bs{W}$:
\begin{equation}
\label{eq_W_balanced}
    \bs{W(t)}= \bs{U}  \bs{\Omega(t)} \bs{\Theta}_{w}\bs{(t)}\bs{Q}
\end{equation}
where $\bs{\Omega}(t) $ is a $p \times n^2$ matrix defined by: 
\begin{equation}
    \begin{cases}
\bs{\Omega}_{\alpha, j} = |\bs{(Q k(t))}_j|,   \, \,  \bs{\Omega}_{\alpha, j_{symm}} = |\bs{(Q k(t))}_j| &  j \rightarrow \alpha  \\
\bs{\Omega}_{\alpha, j} = 0  & else  \\
\end{cases}
\label{eq:W_bar_sol_real}
\end{equation} 
and 
\begin{equation}
     \bs{\Theta}_{w}(t)  = diag(\bs{e}^{i\bs{\delta}_w \bs{(t)}}) 
\end{equation}
With, for all $j \in \sigma_{symm}^{\alpha}$: %$j \in \{0, \cdots, n^2 -1 \}$:
\begin{align}\label{eq:phase_aligned}
& \,(\bs{\delta}_k)_j (t) + (\bs{\delta}_w)_j (t) = - (\bs{\delta}_\phi)_j (t)
\end{align}.

Naturally, the product $\bs{\ol{W}} \bs{\ol{dbc(K)}} \bs{\ol{\Sigma^{xx}}} = \bs{A}$ then becomes: 

\begin{align}
\label{eq_A_balanced}
\begin{cases}
    A_{\alpha,\alpha}(t) = n \sum\limits_{j \in \sigma^{\alpha}_{symm}} |(\bs{Q} \bs{\phi}^\alpha)_j| \, \, |(\bs{Qk(t)})_j|^2 \, \, \ol{\Sigma}_{\alpha,\alpha} & \alpha \in \{ 0, \cdots, p-1 \}\\
     A_{\alpha,\beta}(t) = 0 & \text{else}
\end{cases}
\end{align}

And we also have (cfr. Eq. \ref{eq:transform_W}):
\begin{align}
    & \bs{R} = \bs{Q}^{-1}\\
   & \bs{\ol{W}}(t) = \bs{\Omega}(t) \bs{\Theta_w}(t)\\
   & \bs{W} = \bs{U} \bs{\ol{W}}(t) \bs{Q}
\end{align}

as soon as the parameters become aligned and balanced. Given that this happens fast, the dynamics can be approximated by assuming this shape for $\bs{W}$ from the start.\\

We can plug this in the differential equation given by Eq. \ref{eq:update_diag_Qk_transformed} to obtain:
\begin{align}
&\frac{1}{n \lambda}  \big ( \frac{d \, diag(\bs{Q^{-1}  k})}{dt}  \big )_{j,j} = \bs Q^{-1} _{j,:} \bs{R}^{-*T} \bs{\ol{W}} ^{*T}   \big( \bs{S} - \bs{A}     )  (\bs{Q}\bs{V})^T_{:,j} \\
\iff & \frac{1}{n \lambda}  \big ( \frac{d \, diag(\bs{Q k})}{dt}  \big )_{j,j} = \bs Q_{j,:} \bs{R}^{-*T} \bs{\ol{W}} ^{*T}   \big( \bs{S} - \bs{A}     )  (\bs{Q}^{-1} \bs{V})^T_{:,j} \\
\iff & \frac{1}{n \lambda}  \big ( \frac{d \, diag(\bs{Q k})}{dt}  \big )_{j,j} = \bs Q_{j,:} \bs{Q}^{-1} (\bs{\Omega} \bs{\Theta_w})^{*T}   \big( \bs{S} - \bs{A}     )  (\bs{Q}^{-1} \bs{V})^T_{:,j} \\
\iff & \frac{1}{n \lambda}  \big ( \frac{d \, diag(\bs{Q k})}{dt}  \big )_{j,j} = \bs Q_{j,:} \bs{Q}^{-1} (\bs{\Omega} \bs{\Theta_w})^{*T}   \big( \bs{S} - \bs{A}     )  (\bs{Q}^{-1} \bs{V})^T_{:,j} \\
\iff & \frac{1}{n \lambda}  \big ( \frac{d \, diag(\bs{Q k})}{dt}  \big )_{j,j} = \bs{I}_{j,:} \bs{\Theta_w}^* (\bs{\Omega} )^{T}   \big( \bs{S} - \bs{A}     )  (\bs{Q}^{-1} \bs{V})^T_{:,j} \\
\iff & \frac{1}{n \lambda}  \big ( \frac{d \, diag(\bs{Q k})}{dt}  \big )_{j,j} =\bs{\Theta_w}^*_{j,j} (\bs{\Omega} )^{T}_{j,\alpha}    \big( \bs{S} - \bs{A}     )_{\alpha, \alpha}  (\bs{Q}^{-1} \bs{V})^T_{\alpha,j} \\
\iff & \frac{1}{n \lambda}  \frac{ d \big (  \, |(\bs{Qk})_j| e^{i(\bs{\delta_k})_j} \big )  }{dt}  = e^{-i(\bs{\delta_w})_j} |(\bs{Qk})_j|   \big( \bs{S} - \bs{A}     )_{\alpha, \alpha} e^{-i(\bs{\delta_\phi})_j}  |(\bs{Q}  \bs{\phi}^{\alpha} )_{j}| \\
\iff & \frac{1}{n \lambda}  \frac{ d \big (  \, |(\bs{Qk})_j| e^{i(\bs{\delta_k})_j} \big )  }{dt}  |(\bs{Qk})_j| e^{-i(\bs{\delta_k})_j} = e^{-i ( (\bs{\delta_w})_j + (\bs{\delta_k})_j +(\bs{\delta_\phi})_j )} |(\bs{Qk})_j|^2   \big( \bs{S} - \bs{A}     )_{\alpha, \alpha}  |(\bs{Q}  \bs{\phi}^{\alpha} )_{j}| \\
\iff & \frac{1}{ 2 n \lambda}  \big ( \frac{d \,|(\bs{Qk})_j|^2}{dt}  \big )_{j,j} =  |(\bs{Qk})_j|^2
    ( \bs{S} - \bs{A})_{\alpha,\alpha} | (\bs{Q}  \bs{\phi}^{\alpha} )_{j}  |.
        \label{eq:diag_winner_takes_all}
\end{align}
This is Eq. \ref{eq:diag_winner_takes_all_main} of the main paper, describing the winner-takes-all dynamics.

\section{Analytical Solutions to the Dynamics of Learning for Pure Cosines Datasets} \label{sup:analytical_solutions}

We derive exact, analytical solutions to the evolution of the effective singular values given the following input:
\begin{equation}
    X^{(c)}_{l,m} =  b^{(c)} \cos( 2 \pi \frac{\mu l}{n} +   2 \pi \frac{\nu m}{n})
\end{equation}
as discussed in the main paper. We thus use a \emph{single} cosine per class image, and we set  phase $\delta_j = 0$ (cfr. Eq. \ref{eq:sums_of_cosines}). The choice of the phase does not influence the conclusions, but makes the discussion in the main paper less complicated.

In this case the right singular vectors $\bs{\phi}^{\alpha} = \bs{\phi}^{(\alpha)*} $ are given by:
\begin{align}
 \bs{\phi}^{\alpha}_i &= \frac{1}{d_\alpha n}   \cos( 2 \pi \frac{\mu l}{n} +   2 \pi \frac{\nu m}{n} + (\delta_\phi)_j), 
\end{align}
with $(\delta_\phi)_j= 0$ or $(\delta_\phi)_j= \pi$, introducing a possible minus sign (see Eq. \ref{eq:sums_of_cosines_phi} and discussion following the equation). Here $\mu = div(j,n), \, \nu = mod(j,n)$ are the frequency indices and  $l = div(i,n), \, m = mod(i,n)$ are the pixel indices. For ease of notation, we have introduced $d_{\alpha}$:

\begin{equation}
\label{eq_sup:sol_W}
\begin{cases}
d_{\alpha} = 1 & j = 0,  j \rightarrow \alpha \\
d_{\alpha} = \frac{1}{\sqrt{2}} & j \neq 0,  j \rightarrow \alpha
\end{cases}
\end{equation}
where with $j \rightarrow \alpha$ we indicate that $j$ is the vec-2D index that indexes the horizontal (index $\mu$) and vertical  (index $\nu$)  frequencies that are used to construct the singular vector indexed by $\alpha$. We then find:
\begin{equation}
\begin{cases}
(\bs{Q}  \bs{\phi}^{\alpha} )_{j} =    d_\alpha e^{i (\delta_\phi)_j} &  j \rightarrow \alpha   \\
(\bs{Q}  \bs{\phi}^{\alpha} )_{j_{symm}} =   d_\alpha e^{i (\delta_\phi)_{j_{symm}}}=d_\alpha e^{-i (\delta_\phi)_j} &  j \rightarrow \alpha   \\
(\bs{Q}  \bs{\phi}^{\alpha} )_{j}  = 0  & else  
\end{cases}
\label{eq:VQ_real_equal}
\end{equation}
where we denote with $j_{symm}$ the frequency index that corresponds to the pair of frequency indices $(n-\nu , n- \mu)$. \\
% We now define a type of initial conditions for the fully connected layer we will call the `balanced' initial conditions. Usually, the network is initialised with small, random initial conditions for both the convolutional and the fully connected layers. The balanced initial conditions are instead a type of initial conditions where the kernel is still initialized with small, random values, but the fully connected layer is subsequently initialized with: 
% \begin{equation}
%     \bs{\ol{W}} =  \bs{\Omega} \bs{Q}^{-1} \bs{R},
% \end{equation}
% with $\bs{\Omega}$ a $p \times n^2$ matrix defined by:
% \begin{equation}
%     \begin{cases}
% (\bs{\Omega})_{\alpha, j} = \pm diag(\bs{Q k})_{j,j}  = \pm \bs{(Q k)}_j &  j \rightarrow \alpha  \\
% (\bs{\Omega})_{\alpha, j} = \pm diag(\bs{Q k})^*_{j,j}  = \pm \bs{(Q k)^*}_j & j_{symm} \rightarrow \alpha  \\
% (\bs{\Omega})_{\alpha, j} = 0  & else  \\
% \end{cases}
% \label{eq:W_bar_sol_real}
% \end{equation} 
% such that we have for the now $p \times n^2$ rectangular diagonal product $\bs{\ol{W}} \ol{\bs{dbc(K)}}$:
% \begin{equation}
%     \begin{cases}
% \big(\bs{\ol{W}} \ol{\bs{dbc(K)}}\big)_{\alpha, \beta} = \frac{1}{d_\alpha^2} n d_\alpha |\bs{(Q k)}_j|^2 &  j \rightarrow \alpha, \alpha = \beta, \alpha \in \{0,p-1\} \\
% \big(\bs{\ol{W}} \ol{\bs{dbc(K)}}\big)_{\alpha, \beta}  = 0  & else  \\
% \end{cases}
% \end{equation}
% where the factor $\frac{1}{d_\alpha^2}=2$ for non-zero frequencies, and $\frac{1}{d_\alpha^2}=1$  for the zero frequency (this thus relates to the symmetries in the spectrum). 

We initialize the kernel with small, random weights, and set the initial weights of the fully connected layer according to Eq. \ref{eq_W_balanced}. This makes the network balanced and aligned. We get (see Eq. \ref{eq_A_balanced}):

\begin{align}
\begin{cases}
    A_{\alpha,\alpha}(t) = 2 n d_\alpha  |(\bs{Qk(t)})_j|^2 \, \, \ol{\Sigma}_{\alpha,\alpha} & j \rightarrow \alpha,  j \neq 0, \alpha \in \{ 0, \cdots, p-1 \}\\
    A_{\alpha,\alpha}(t) = n d_\alpha  |(\bs{Qk(t)})_j|^2 \, \, \ol{\Sigma}_{\alpha,\alpha} & j \rightarrow \alpha, j = 0, \alpha \in \{ 0, \cdots, p-1 \}\\
     A_{\alpha,\beta}(t) = 0 & \text{else}
\end{cases}
\end{align}
Which can be succinctly written as:
\begin{equation}
    \bs{A}_{\alpha, \alpha} = \frac{n}{d_\alpha }|(\bs{Q k})_j|^2 \ol{\bs{\Sigma}^{ x x}}_{\alpha, \alpha}
\end{equation}
and $\bs{A}_{\alpha, \beta} = 0$  if $\alpha \neq \beta$.

These initial conditions decouple the general system of equations (eqs. \ref{eq:update_diag_Qk_transformed} and \ref{eq:update_W_bar}); we can use Eq. \ref{eq:diag_winner_takes_all} 
for the specific pairs of modes and frequencies $(\alpha, j)$ used in the single cosines dataset; all other derivatives are zero. From this equation and the fact that we only have one frequency $j$ per mode $\alpha$, we can derive the full evolution of the effective singular values $A_{\alpha, \alpha}$:
\begin{align}
    \frac{d  \bs{A}_{\alpha, \alpha}}{dt} &=  \frac{n}{d_\alpha} \frac{d  |(\bs{Q k})_j|^2}{dt} \ol{\bs{\Sigma}^{ x x}}_{\alpha, \alpha} \\
   &= 2 n \lambda  \frac{n}{d_\alpha}  \ol{\bs{\Sigma}^{ x x}}_{\alpha, \alpha} |(\bs{Q k})_j|^2 ( \bs{S} - \bs{A}     )_{\alpha,\alpha} d_{\alpha}\\
    &= 2 n d_\alpha \lambda \bs{A}_{\alpha,\alpha}  ( \bs{S}_{\alpha,\alpha} - \bs{A}_{\alpha,\alpha})
\end{align}

This is a non-linear, separable differential equation that we can solve for time $t$ (see also \cite{saxe2019mathematical}). To simplify notation, we will use $a_\alpha = A_{\alpha, \alpha}$ and $s_\alpha = S_{\alpha, \alpha}$. Using 
\begin{equation}
    t = \frac{1}{2 n d_\alpha \lambda} \int_{a^{(0)}_\alpha}^{a_\alpha^{(f)}} \frac{da_\alpha}{a_\alpha(s_\alpha-a_\alpha)} =  \frac{1}{2 n d_\alpha \lambda s_{\alpha}} ln \Big(\frac{a_\alpha^{(f)} (s_\alpha - a^{(0)}_\alpha) }{a_\alpha^{(0)} (s_\alpha - a^{(f)}_\alpha)} \Big)
\end{equation}
where $t$ is the training time (in number of samples) required to arrive at value $a_\alpha^{(f)}$ at time t from initial value $a_\alpha^{(0)}$. The evolution during training  of the effective singular value, $a_{\alpha}(t)$, is then given by:
\begin{equation}
    a_{\alpha} (t) = \frac{s_{\alpha} e^{ 2  \lambda n  d_\alpha s_{\alpha} t}}{ e^{ 2 \lambda d_\alpha n s_{\alpha} t} - 1 + s_{\alpha} / a_{\alpha}^{(0)} }.
    \label{eq:a_complex_waves}
\end{equation}

\section{Details of experiments} \label{sec_sup:experiments}

All experiments are implemented with tensorflow/keras. A factor 1/p is used in the implementation of the MSE loss, with p the number of classes. All experiments can be completed in a couple of hours on a basic CPU. 

When only one image per class is used, this means in practice that for each update, a sample/class is randomly selected. For the CIFAR-4 dataset, we used the more conventional setup with a number of epochs, wherein the samples of the train set are given as input in a random order. 

We use CNNs with a single convolutional layer, a flatten layer, and a single fully connected layer. There are only linear activation functions. The convolutional layer consists of a single kernel $\bs{K}$ with the same dimensions as the input images ($n \times n$) for grayscale images, and 3 such kernels for colour images. Images are circularly padded beforehand: we extend the images with n-1 rows/columns of the pixel values of the opposite side. When the kernel of $n\times n$ is applied, the padding is then in practice equivalent to wrapping the kernel around the image when it slides over the original image boundary. The result of this convolution is again an $n \times n$ matrix.\

FCNNs consist of two fully connected layers; the hidden one with $n^2$ nodes, the output one with p nodes. The first weight matrix therefore has dimensions $n^2 \times n$, the second $p \times n^2$. The choice of nodes in the hidden layer corresponds to the architecture of the CNN, which performs a convolution from an $n \times n$ to an $n \times n$ image, and therefore has an associated doubly block circulant matrix of size $n^2 \times n^2$. For the CIFAR-4 dataset, we use $3n^2$ nodes in the hidden layer. Images are flattened to $n^2 \times 1$ or $3n^2 \times 1$ vectors beforehand.

\subsection{Pure cosines experiment}
\begin{figure}[h!]
   \centering
   \includegraphics[width=0.4\linewidth]{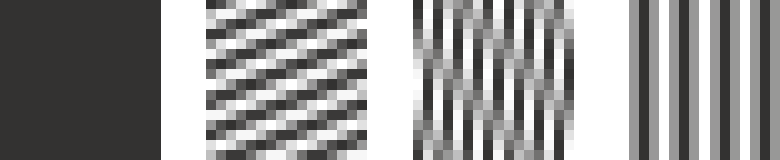}
   %\captionof{figure}{Figure caption}
   \caption{Cosines used in the `pure cosines' experiment. }
   \label{fig:pure_cosines}
 \end{figure}

Image dimension: $16 \times 16$; horizontal and vertical freq. pairs for each class: (0,0), (5,2), (1,7), (0,4); amplitudes for each class: 1.5, 1, 0.5, 0.2.

\begin{itemize}
    \item $\lambda_{CNN} = 1/2000$, $\lambda_{FCNN} = 16/2000$
    \item init CNN: random normal with $\mu=0$ and $\sigma= 0.00001$ for the kernel $\bs{K}$, weights $\bs{W}$ balanced to $\bs{K}$ using $\bs{R} = \bs{Q}$ (eq. \ref{eq_sup:sol_W}). This makes $\bs{A}_{init}$ rectangular diagonal. 
    \item init FCNN: using $\bs{A}_{init}$ from CNN, setting the first p diagonal elements of $\bs{\ol{W}^1}$ and $\bs{\ol{W}^2}$ to the value $\sqrt{ \frac{\bs{A}_{(init) \alpha, \alpha}}{\ol{\Sigma^{xx}_\alpha}}}$, other elements to zero. 
    \item number of updates: 8000
    \item number of repeated experiments: 10
\end{itemize}

\subsection{Dominant frequency bias experiment (sums of cosines)}

Sums of cosines dataset as shown in fig. \ref{fig:CNN_dominant_freq}, image dimensions $64 \times 64$.

\begin{itemize}
    \item $\lambda_{CNN} = 1/10000$,$\lambda_{FCNN} = 64/10000$
    \item init CNN: random normal with $\mu=0$ and $\sigma= 0.00001$ for the kernel and weights.
    \item init FCNN: random normal with $\mu=0$ and $\sigma= 0.00001$ for all weights.
    \item number of updates: 600
    \item number of repeated experiments: 10
\end{itemize}

\subsection{Geometric shapes experiment}

Dataset as shown in fig. \ref{fig:geometric_dataset_structure}. Image dimensions  $64 \times 64$.

\begin{itemize}
    \item $\lambda_{CNN} = 1/20000$, $\lambda_{FCNN} = 64/20000$
    \item init CNN: random normal with $\mu=0$ and $\sigma= 0.00001$ for the kernel and weights.
    \item init FCNN: random normal with $\mu=0$ and $\sigma= 0.00001$ for all weights.
    \item number of updates: 60000
    \item number of repeated experiments: 10
\end{itemize}

\subsection{CIFAR-4 experiment} 

 \begin{figure}[h!]
   \centering
   \includegraphics[width=0.4\linewidth]{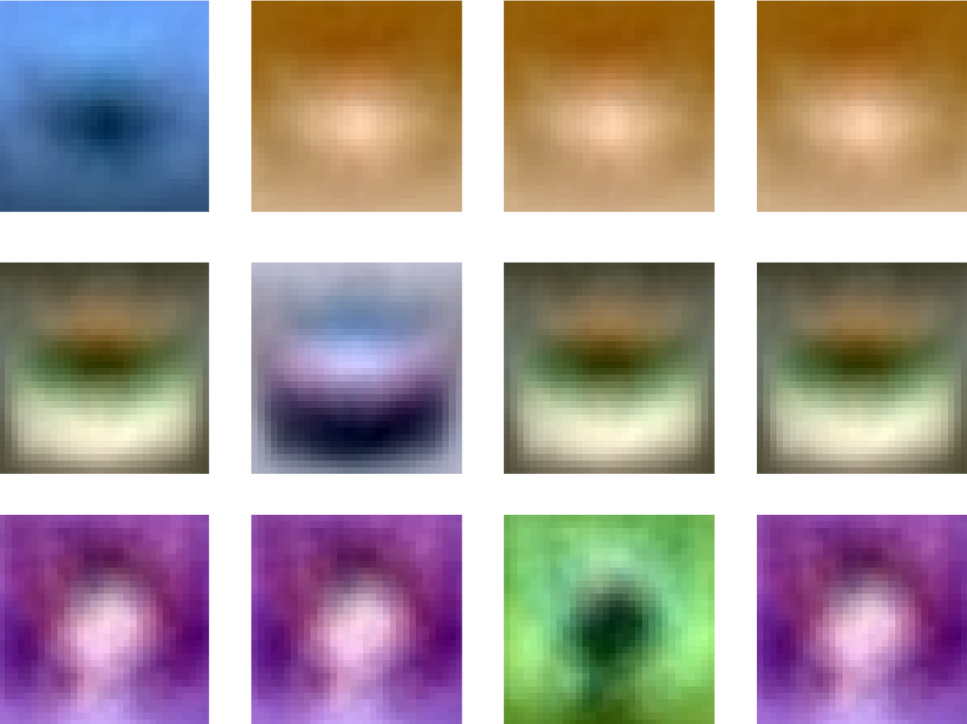}
   %\captionof{figure}{Figure caption}
   \caption{Modes of the CIFAR-4 dataset with removed first mode (i.e., subtracted mean over samples). Three modes remain to distinguish between the four classes. The top row is a visualisation of the first mode, the second row of the second mode, etc. $1 \times 3n^2$ vectors are visualised as $n \times n$ color images. Each image is normalised to values between 0 and 1. In this way, we can see that the first mode distinguishes central objects in a blue background, which is especially relevant for the first class `airplane'; the second mode discovers the class `automobile' based on a reddish color and a vague frontal view of a car, and the third mode distinguishes `bird' based in a central object in a predominantly green background. }
   \label{fig:modes_cifar_4}
 \end{figure}

First four classes of the CIFAR-10 dataset. Train set consists of the 5000  samples for each class in the CIFAR-10 train set (thus 20000 samples in total) with mean subtracted, test set consist of the 1000 samples for each class in the CIFAR-10 test set (4000 samples in total) with mean subtracted. Image dimension $32 \times 32$.

\begin{itemize}
    \item $\lambda_{CNN} = 1/10000$, $\lambda_{FCNN} = 32/10000$
    \item init CNN: random normal with $\mu=0$ and $\sigma= 0.0001$ for the kernel and weights.
    \item init FCNN: random normal with $\mu=0$ and $\sigma= 0.0001$ for all weights.
    \item number of updates: 100000
    \item number of repeated experiments: 10
\end{itemize}

\section{Comparison of the 2D-vec Fourier Spectra of the Kernel and the Singular Vectors} \label{sec_sup:compare_spectra}

\begin{figure}[h!]
   \centering
   \includegraphics[width=\textwidth]{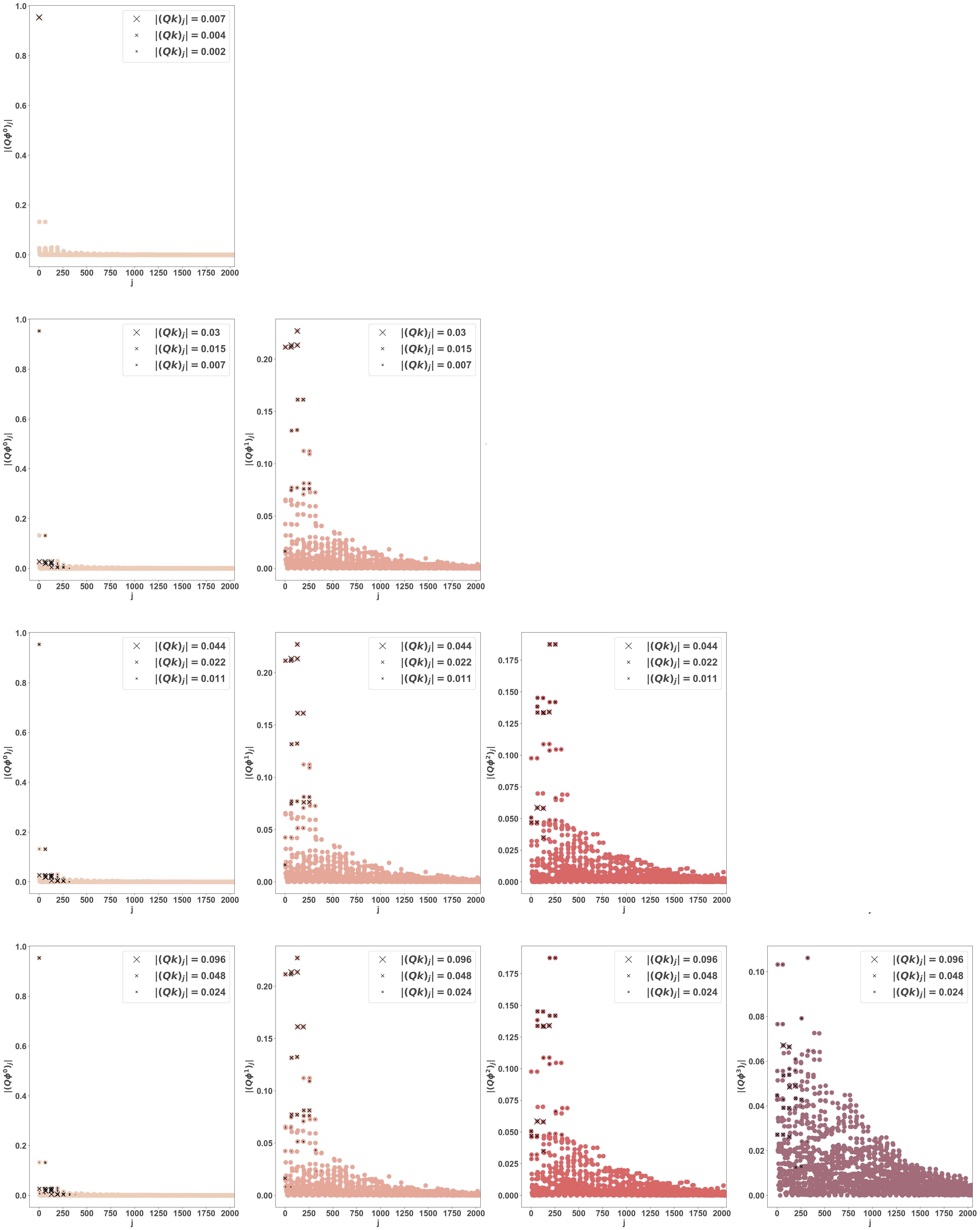}
   %\captionof{figure}{Figure caption}
   \caption{Comparison of the 2D-vec Fourier Spectra of the Kernel and the Singular Vectors at Selected Timepoints.  }
   \label{fig_sup:comparison_of_spectra}
 \end{figure}

In Fig. \ref{fig_sup:comparison_of_spectra} (next page), we visualize that the frequencies that make up the kernel are the dominant frequencies of the singular vectors. The data shown are from the experiment with the geometric shapes dataset  (cfr. Fig. \ref{fig:results_experiments}). 

Each row of the figure corresonds to a timepoint during training with the CNN:

\begin{itemize}
    \item row 0, timepoint = 2000 samples, mode 0 discovered; 
    \item row 1, timepoint = 9000 samples, mode 1 discovered; 
    \item  row 2, timepoint = 16000 samples, mode 2 discovered;
    \item row 3, timepoint = 55000 samples, mode 3 discovered 
\end{itemize}   
(cfr. Fig. \ref{fig:results_experiments} (a)). At each row we plot the 2D-vec Fourier spectra of the singular vectors, i.e., the values $ |(\bs{Q} \bs{\phi}^{\alpha})_j|$, that correspond to the modes discovered at that point. These spectra are unaltered during training, and are thus repeated across rows. The kernel, and therefore the values $ |(\bs{Q} \bs{k})_j|$, do change during training. We consider the average of the values $ |(\bs{Q} \bs{k})_j|$ over the number of experiment runs. We overlay these averaged  values $ |(\bs{Q} \bs{k})_j|$  for the given timepoints at the corresponding values $ |(\bs{Q} \bs{\phi}^{\alpha})_j|$. This  overlay is indicated with black crosses. The size of these crosses codes for the magnitude of the values $ |(\bs{Q} \bs{k})_j|$. In this way, we can first note that the Fourier spectrum of $k$ is much sparser than the different Fourier spectra of the singular vectors (in the spectrum of each singular vectors, the number of crosses is much smaller than the total number of other markers). Moreover, large values $ |(\bs{Q} \bs{k})_j|$ (large crosses) mainly correspond to large values of  $ |(\bs{Q} \bs{\phi}^{\alpha})_j|$. Thus the kernel is, at any point during training, a sparse mixture of the dominant frequencies of the singular vectors discovered so far.

\section{Details of Experiments with Deep Networks}\label{sec_sup:deep_networks}
The deep networks have the following architecture:
\begin{itemize}
    \item a convolutional layer, 16 channels, zero padding
    \item a convolutional layer, 16 channels, zero padding
    \item (non-linear network) a max-pooling layer with size $2\times 2$
    \item a convolutional layer, 16 channels, zero padding
    \item a convolutional layer, 16 channels, zero padding
    \item (non-linear network) a max-pooling layer with size $2\times 2$
    \item flatten layer
    \item fully connected layer with 16 nodes 
    \item fully connected layer to 10 output nodes 
\end{itemize}

For the non-linear network, we apply the non-linear ReLU activation functions on all convolutional layers and the first fully connected layer. The last layer has no activation function (=linear activation).\\

We use the MSE loss and the full dataset, i.e., all 50000 samples in the training set. The batch size is set to 8 and there are thus 6240 batches per epoch. We train the networks for 25 epochs, yielding a total of 156000 batches. Every 500 batches we compute the test and train loss on a subset of 5000 train and test samples, respectively. At these points, we also compute $\bs{\Sigma^{\hat y  x}}$ (eq. \ref{eq:sigma_y_hat_x}) and, from this result, we compute the matrix $\bs{A}$  (eq. \ref{eq:SVD_sigma_y_hat_x}). \\

The network is initialized with the Glorot normal initializer. This is a truncated normal distribution with mean 0 and standard deviation $\sqrt{2 / (\textnormal{fan in} + \textnormal{fan out})}$.  Here \emph{fan in} is the number of input units in the weight matrix and \emph{fan out} is the number of output units in the weight matrix. We scaled the values drawn from this initializer with a factor $0.1$ to ensure that the weights would be small enough to yield a rich training regime. \\

\end{document}